\documentclass[lettersize,journal]{IEEEtran}
\usepackage[caption=false,subrefformat=parens]{subfig} 
\usepackage{cite}
\usepackage{algorithm}
\usepackage{amsmath,amssymb,amsfonts}
\usepackage[noend]{algpseudocode}
\usepackage{graphicx}
\usepackage{textcomp}
\usepackage{xcolor}
\usepackage{float}
\usepackage{picinpar}
\usepackage{babel}
\usepackage{booktabs}
\usepackage{multirow}
\usepackage{blindtext}
\usepackage{balance}
\usepackage{hyperref}
\usepackage{bbm}
\usepackage{subfig}
\usepackage{stfloats}
\usepackage{mathtools}
\usepackage{amsthm}
\usepackage{color}
\hypersetup{hidelinks}

\usepackage{listings}
\usepackage[utf8]{inputenc}
\usepackage{amsmath}
\usepackage{listings}
\usepackage{enumitem}
\lstdefinelanguage{json}{
    basicstyle=\ttfamily\small,
    commentstyle=\color{gray},
    stringstyle=\color{blue},
    numberstyle=\tiny\color{gray},
    keywordstyle=\color{purple},
    breaklines=true,
    showstringspaces=false,
    frame=none,
    rulecolor=\color{lightgray}
}

\newcommand{\jsoninbox}[1]{
    \begin{lstlisting}[language=json, basicstyle=\ttfamily\fontsize{9}{10}\selectfont]
#1
    \end{lstlisting}
}

\newcommand{\promtblock}[1]{\vspace{1pt}\noindent\fbox{\parbox{0.99\linewidth}{{\fontsize{9}{10} \selectfont  \textcolor{black} {{#1}}}}}\vspace{1pt}}
\newcommand{\llmoutput}[1]{\fcolorbox{light-gray}{light-gray}{\parbox{0.98\linewidth}{{\fontsize{9}{10} \selectfont  \textcolor{black}{{#1}}}}}}

\definecolor{light-gray}{HTML}{F2F2F2}
\definecolor{darkblue}{RGB}{0,20,115}
\definecolor{purple}{RGB}{112,48,160}
\definecolor{blue}{RGB}{20,50,190}

\usepackage{amsmath,amsfonts,bm}

\def\eqref#1{equation~\ref{#1}}

\def\1{\bm{1}}

\def\eps{{\epsilon}}

\def\rva{{\mathbf{a}}}

\def\rvo{{\mathbf{o}}}

\def\rvx{{\mathbf{x}}}
\def\rvy{{\mathbf{y}}}
\def\rvz{{\mathbf{z}}}

\def\rmI{{\mathbf{I}}}

\def\vzero{{\bm{0}}}

\def\va{{\bm{a}}}

\def\vf{{\bm{f}}}
\def\vg{{\bm{g}}}

\def\vk{{\bm{k}}}

\def\vo{{\bm{o}}}

\def\vq{{\bm{q}}}

\def\vx{{\bm{x}}}
\def\vy{{\bm{y}}}

\def\evm{{m}}

\def\mF{{\bm{F}}}

\def\mI{{\bm{I}}}

\DeclareMathAlphabet{\mathsfit}{\encodingdefault}{\sfdefault}{m}{sl}
\SetMathAlphabet{\mathsfit}{bold}{\encodingdefault}{\sfdefault}{bx}{n}

\def\gD{{\mathcal{D}}}

\def\gK{{\mathcal{K}}}
\def\gL{{\mathcal{L}}}

\def\gN{{\mathcal{N}}}

\def\gT{{\mathcal{T}}}
\def\gU{{\mathcal{U}}}

\newcommand{\pdata}{p_{\rm{data}}}

\newcommand{\E}{\mathbb{E}}

\newcommand*{\dif}{\mathop{}\!\mathrm{d}}

\usepackage{tikz}
\definecolor{lime}{HTML}{A6CE39}
\DeclareRobustCommand{\orcidicon}{%
    \begin{tikzpicture}
        \draw[lime, fill=lime] (0,0) 
        circle [radius=0.16] 
        node[white] {{\fontfamily{qag}\selectfont \tiny ID}};
        \draw[white, fill=white] (-0.0625,0.095) 
        circle [radius=0.007];    
    \end{tikzpicture}
    \hspace{-2mm}
}
\foreach \x in {A, ..., Z}{%
    \expandafter\xdef\csname orcid\x\endcsname{\noexpand\href{https://orcid.org/\csname orcidauthor\x\endcsname}{\noexpand\orcidicon}}
}

\begin{document} 

\title{BiKC+: Bimanual Hierarchical Imitation with Keypose-Conditioned Coordination-Aware Consistency Policies}

\author{
    Hang Xu \orcidA{},
    Yizhou Chen$^{\dag}$ \orcidB{},
    Dongjie Yu \orcidC{},
    Yi Ren \orcidD{} and
    Jia Pan$^{\dag}$ \orcidE{}
    \thanks{
        This study is partially supported by Jiangsu-Hong Kong-Macau Project BZ2024061, RGC grants (GRF 17201025, GRF 17200924, NSFC-RGC Joint Research Scheme N\_HKU705/24), and the Natural Science Foundation of China (Project Number 62461160309).
        (\textit{$\dag$Correspondence authors: Jia Pan and Yizhou Chen.})
        }
    \thanks{
        Hang Xu is with the Centre for Transformative Garment Production, The University of Hong Kong, Hong Kong, China. (e-mail: {xuhang\_official@outlook.com}).
    }
    \thanks{
        Yizhou Chen, Dongjie Yu and Jia Pan are with the School of Computing and Data Science, The University of Hong Kong, Hong Kong, China. (e-mail: {yzhchen@hku.hk}; {djyu@connect.hku.hk}; {jpan@cs.hku.hk}).
    }
    \thanks{
        Yi Ren is with Advanced Manufacturing Lab, Huawei Technologies, Shenzhen, Guangdong, China. (e-mail: {even.renyi@huawei.com}).
    }
}

\markboth{Accepted by IEEE Transactions on Automation Science and Engineering}%
{Hang Xu \MakeLowercase{\textit{et al.}}:BiKC+: Bimanual Hierarchical Imitation with Keypose-Conditioned Coordination-Aware Consistency Policies}


\maketitle

\begin{abstract} 

Robots are essential in industrial manufacturing due to their reliability and efficiency. They excel in performing simple and repetitive unimanual tasks but still face challenges with bimanual manipulation. This difficulty arises from the complexities of coordinating dual arms and handling multi-stage processes.
Recent integration of generative models into imitation learning (IL) has made progress in tackling specific challenges. However, few approaches explicitly consider the multi-stage nature of bimanual tasks while also emphasizing the importance of inference speed. In multi-stage tasks, failures or delays at any stage can cascade over time, impacting the success and efficiency of subsequent sub-stages and ultimately hindering overall task performance.
In this paper, we propose a novel keypose-conditioned coordination-aware consistency policy tailored for bimanual manipulation. Our framework instantiates hierarchical imitation learning with a high-level keypose predictor and a low-level trajectory generator. 
The predicted keyposes serve as sub-goals for trajectory generation, indicating targets for individual sub-stages. 
The trajectory generator is formulated as a consistency model, generating action sequences based on historical observations and predicted keyposes in a single inference step. 
In particular, we devise an innovative approach for identifying bimanual keyposes, considering both robot-centric action features and task-centric operation styles.
Simulation and real-world experiments illustrate that our approach significantly outperforms baseline methods in terms of success rates and operational efficiency.
Implementation codes can be found at \textit{\url{https://github.com/JoanaHXU/BiKC-plus}}.

\end{abstract}

\def\abstractname{Note to Practitioners} 
\begin{abstract}
Bimanual manipulation typically involves multiple stages that require efficient interactions between two arms, presenting both step-wise and stage-wise challenges for imitation learning (IL) systems. Existing approaches, particularly those based on generative models, have not explicitly considered the impact of these multiple stages, which negatively affects overall success rates and operational efficiency. 
This paper proposes a novel hierarchical IL framework, BiKC+, which learns from distributionally multi-modal demonstrations, generates actions through one-step inference, and addresses bimanual multi-stage manipulation. 
The high-level keypose predictor forecasts the next target keypose in joint space, serving as a guidance for low-level actions and an indicator for sub-stage completion. 
This enhances per-stage reliability and improves overall success rates. The low-level trajectory predictor is formulated as a consistency model that generates action sequences in a single inference step, thereby increasing inference speed and enhancing operational efficiency.
Comprehensive experiments indicate the potential of the proposed framework to be applied in real-world manufacturing and industrial settings.
Future research will integrate additional sensory information, such as forces and torques, to accurately reproduce subtle motions arising from contact and force interactions, thereby strengthening the approach for fine-grained manipulation.
\end{abstract}


\section{Introduction} 
\label{sec:intro}

\IEEEPARstart{R}{obots} have become indispensable in industrial manufacturing, significantly enhancing productivity across a wide range of fields~\cite{sun2023integrating, zhang2024oneshota}. They particularly excel in performing simple and repetitive tasks, such as picking, packing, and placing~\cite{xiong2024enhancing, mehrotra2023innovation}. However, deploying robots in bimanual multi-stage manipulation tasks, such as cloth folding, battery slotting, and tool assembling, remains limited~\cite{chi2024universal, zhao2023learning, grotz2024peract2, bahety2024screwmimic}. These tasks require precise dual-arm coordination across multiple stages, making them more complex than unimanual tasks~\cite{gao2024bikvil, varley2024embodied}.
A conventional solution is to assign a specific role to each arm (e.g., leader and follower) based on human-specified rules, with each arm programmed independently~\cite{krebs2022bimanual, grannen2023stabilize, liu2024voxactb, bahety2024screwmimic}. However, this approach is labor-intensive and hard-coded, as coordination rules must be customized for each specific task, especially those involving multiple stages~\cite{liu2022robot}.

Recently, Imitation Learning (IL) has been proposed to address bimanual manipulation. This straightforward yet powerful method enables the development of automated agents in an end-to-end fashion by imitating provided demonstrations, eliminating the need for explicitly and manually crafted coordination rules~\cite{ding2019goalconditioned, xie2020deep, kujanpaa2023hierarchical}.

Nevertheless, applying IL to bimanual manipulation faces significant challenges due to the multi-stage nature: per-stage reliability and per-step efficiency.
First, the performance (i.e., success or failure) of a previous stage has a lasting impact on subsequent stages. For example, when transferring a cube, if the robot fails to pick it up, it will inevitably fail to pass it on. Since failure can occur at any stage, multi-stage tasks are more susceptible to failure compared to single-stage ones.
Second, multi-stage tasks typically have longer horizons, leading to accumulated latency and consequently extending the overall task duration. This clearly harms efficiency, particularly in real-world systems. Furthermore, in the context of bimanual tasks, demonstrations tend to be highly diverse due to increased degrees of freedom (DOFs)~\cite{krebs2022bimanual}. 
This diversity results in distributionally multi-modal demonstrations, characterized by different behavioral styles or stage orders, which present significant challenges for accurate imitation.
Therefore, addressing these challenges is essential for applying robots to bimanual manipulation tasks in real world.

Recently, advanced generative models have been integrated into imitation learning, making a significant breakthrough in addressing the challenges in bimanual manipulation tasks.
For instance, Action Chunking with Transformers (ACT) ~\cite{zhao2023learning} formulates the robot policy as a conditioned variational autoencoder (cVAE), while Diffusion Policy (DP)~\cite{chi2023diffusion} employs denoising diffusion probabilistic models (DDPMs) for policy formulation. Both approaches reduce compounding errors by generating action sequences and tackling distributionally multi-modal demonstrations. However, ACT is constrained by a fixed behavior style during evaluation, and DP is limited in inference speed. Both overlook the inevitable sub-goals necessary for whole-task completion, which diminishes overall success in multi-stage tasks.
An insight into how humans tackle multi-stage tasks reveals that humans generally predict keyposes (i.e., upcoming target poses) and view them as sub-goals. This subgoal helps to direct immediate actions over a shorter period~\cite{nakahashi2016modeling, binder2023humans}. By doing so, they effectively break down a complex task into several more manageable sub-stages. Similar concepts have been explored in unimanual manipulation~\cite{ma2024hierarchical}. By predicting the next keypose and connecting keyposes with continuous trajectories derived from either motion planning~\cite{gervet2023act3d} or learned planners~\cite{xian2023chaineddiffuser}, these methods have shown success in long-horizon tasks. While keypose-based methods have proven effective, they have not been tailored or evaluated for bimanual scenarios due to the intricate spatio-temporal coordination relationships between two arms.

To this end, this paper proposes \underline{Bi}manual \underline{K}eypose-conditioned \underline{C}oordination-aware \underline{C}onsistency Policy (BiKC+). 
This hierarchical IL framework overcomes limitations of existing methods by learning from distributionally multi-modal demonstrations and generates actions with one-step inference, enabling reliable and efficient bimanual multi-stage manipulation.
Our work tackles the critical challenge of bimanual keypose identification through a three-stage pipeline: 1) extracting unimanual keyposes using heuristic rules; 2) identifying coordination modes via a proposed vision-language-model (VLM)-assisted contact-aware solution; and 3) constructing bimanual keyposes through a coordination-driven merging strategy.
This pipeline uniquely integrates both robot-centric action features and task-centric operation styles. 
It ensures spatio-temporal synchronization of the two arms during coordination phase, while preserving the flexibility of individual arms in non-coordination modes, laying the foundation for the BiKC+ framework.
The proposed hierarchical IL framework comprises a high-level keypose predictor as a sub-goal planner and a low-level trajectory generator as a behavior policy.
At the high level, the keypose predictor forecasts the next target keypose in joint space. 
The keypose acts as both a guidance for low-level actions and an indicator for sub-stage completion, which enhances per-stage reliability and thereby improves the overall success rate.
The low-level trajectory generator is formulated as a Consistency Model (CM) and trained from scratch through consistency training~\cite{song2023consistency, song2024improved}. It generates a short-horizon action sequence based on observations, including vision and proprioception, as well as the predicted target keypose. The process is similar to DP but requires only one-step inference while maintaining sample quality.
We conducted experiments on BiKC+ in both simulated and real-world environments, demonstrating that BiKC+ can enhance the overall success rates of multi-stage tasks while significantly improving operational efficiency.

Our contributions are three-fold:
\begin{itemize}
    \item We propose a three-stage pipeline for bimanual keypose identification, integrating robot-centric and task-centric features via VLM-assisted contact awareness and coordination-driven merging, to ensure synchronization in coordination and flexibility in non-coordination.
    \item We present the BiKC+ framework, a hierarchical IL architecture with a keypose predictor that forecasts target keyposes, providing low-level action guidance to enhance sub-stage reliability.
    \item We design a scratch-trained Consistency Model (CM) as the trajectory generator, enabling one-step inference of high-quality action sequences to balance efficiency and sample quality.
\end{itemize}

We organize the rest of this paper as follows. 
Section \ref{sec:related} reviews imitation learning-based approaches for multi-stage bimanual robotic manipulation and highlights BiKC+'s contributions to this field. This is followed by an introduction to the preliminaries of Consistency Models in Section \ref{sec:preliminary}.
Section \ref{sec:method} details our proposed BiKC+ framework, including the design of bimanual keypose identification, as well as the formalization of high-level keypose predictor and low-level trajectory generator as consistency models.
Section \ref{sec:experiments} presents empirical evaluations of BiKC+'s performance in both simulated and real-world manipulation tasks.
Finally, Section \ref{sec:conclusion} summarizes the key contributions of this work, and Section \ref{sec:discuss} discusses the limitations and outlines potential avenues for future research.

\section{Related Work} 
\label{sec:related}

\subsection{Bimanual Manipulation with Imitation Learning}

Bimanual manipulation typically requires intricate coordination between both arms, leading to greater complexity compared to unimanual manipulation \cite{gao2024bikvil, varley2024embodied}. 
Various approaches have been developed to tackle the coordination issue. 
One research stream explicitly models spatial-temporal relationships between two arms, thereby restricting the motion primitives each arm can perform to simplify these tasks~\cite{krebs2022bimanual, liu2022robot, gao2024bikvil, bahety2024screwmimic}.
While the explicit modeling approach is interpretable, it has limited task coverage, given the diversity of bimanual tasks.
There may not be a universal rule that describes all types of relationships in bimanual manipulation. Consequently, imitation learning has become increasingly popular as it enables robots to learn skills directly from expert demonstrations.
The standard procedure for imitating unimanual manipulation includes: (1) collecting demonstrations through teleoperation interfaces~\cite{zhao2023learning, chi2024universal, hirao2023body}; and (2) leveraging advanced parameterized models to approximate the demonstrated state-action mappings~\cite{wang2024novel, xu2024leto}.
Despite its straightforwardness and efficiency, imitation learning is susceptible to compounding errors \cite{zhao2023learning} and distributional multi-modality in demonstrations \cite{chi2023diffusion}, which become more pronounced in bimanual scenarios.

As research advances, integrating generative modeling into IL has achieved significant success in addressing challenges related to compounding errors and distributionally multi-modal demonstrations \cite{chi2024universal, zhao2023learning, chi2023diffusion, shi2023waypointbased}. 
These methods effectively reduce compounding errors by predicting sequences of actions, thus reducing numbers of inference. Furthermore, they tackle the multi-modal challenge with the power of approximating complex data distribution from generative models.
For example, ACT~\cite{zhao2023learning} formulates the robot policy as a conditioned variational autoencoder (cVAE). It addresses the multi-modal issue by extracting the primary behavior style from demonstrations, sticking to the main behavior style during evaluation. It also uses temporal ensemble to produce smoother actions, improving the action smoothness while bringing extra computation costs. 
Another impressive IL method, DP~\cite{chi2023diffusion} captures multi-modal action distributions by representing the policy as a diffusion model, which progressively refines action sequences from random noise. 
While this approach is effective in managing multi-modality, the iterative denoising process introduces significant inference latency. This latency is a crucial concern, especially considering applications in real-world scenarios.

In this study, we propose to use a variant of diffusion models, Consistency Models (CMs)~\cite{song2023consistency, song2024improved}, to formulate robot policies. Similar to DP, CMs are capable of capturing distributional multi-modality in demonstrations. They stand out for their fast inference speed by generating action sequences in a single step. 
Concurrent research \cite{lu2024manicm, prasad2024consistency} has also applied CMs to robotic manipulation, primarily focusing on accelerating inference. However, these studies typically rely on distillation from pretrained diffusion models and are limited to unimanual tasks.
To avoid this issue, this paper uses consistency training (CT) \cite{song2023consistency, song2024improved}, which involves training a CM-based policy from scratch. Comprehensive experiments demonstrate its high-quality performance and real-time deployment in bimanual tasks.

\subsection{Hierarchical Imitation for Multi-stage Tasks}

Hierarchical IL (HIL) is introduced to improve performance in multi-stage tasks by breaking down complex tasks into smaller, more manageable components.
One line of studies is based on a set of predefined manipulation primitives, such as picking, placing, moving, and stabilizing. These primitives are constrained by a predefined skill library \cite{triantafyllidis2023hybrid} or are derived from task-agnostic datasets \cite{wan2024lotus, ma2024hierarchical, reuss2023goalconditioned}. 
They are then organized sequentially using a high-level planner, often referred to as a meta-policy, to complete multi-stage tasks. 
In the context of bimanual tasks, this concept is realized by assigning distinct roles to each arm, thereby allowing each arm to independently learn a policy tailored to its specific function~\cite{grannen2023stabilize}. This method effectively decomposes the bimanual system into two unimanual cases; however, its application is limited to certain types of tasks.
Another line of work segments demonstrations into sub-stage trajectories. 
Heuristics~\cite{xian2023chaineddiffuser, ma2024hierarchical} or unsupervised clustering~\cite{zhu2022bottomup} are leveraged to break down long trajectories into sub-stage segments. The points marking these segments are referred to as \textit{keyposes} and act as sub-goals to be achieved. As a result, the overall framework involves modeling the transitions between high-level sub-goals (task plans) and the low-level actions to accomplish these sub-goals.
Built upon this notion, several studies have made progress in unimanual manipulation~\cite{gervet2023act3d,xian2023chaineddiffuser, ma2024hierarchical}. However, applying keypose to bimanual tasks presents challenges in coordinating the two arms.

To solve the challenges regarding bimanual keyposes, BiKC \cite{yu2024bikc} proposes a straightforward method: merging two sets of unimanual keyposes to form bimanual ones, which circumvents the need to understand bimanual coordination. 
While this merging method works effectively for compatible and perfect demonstrations, it is overly restrictive, limiting the flexibility of individual arms' movement. More importantly, it ignores the bimanual operation styles, without considering how the bimanual coordination affects the keypose definition. 
Extended from BiKC, this work (BiKC+) develops a coordination-driven merging strategy, which incorporates the understanding of bimanual coordination via the proposed VLM-assisted contact-aware solution. This approach ensures the dual-arm synchronization during the coordination phase while preserving the flexibility of unimanual execution in non-coordination mode. 
Furthermore, to capture the flexibility inherent in keyposes, BiKC+ formulates the high-level keypose predictor as a Consistency Model (CM), which excels at modeling multi-modality. Thus, the CM-based keypose predictor can handle the predictions for the multi-modal keyposes derived from partially contradicting demonstrations. 

Notably, Bi-KVIL \cite{gao2024bikvil} also employs keyposes in bimanual manipulation to model object-gripper spatial relationships. 
However, its hierarchical architecture diverges fundamentally from conventional hierarchical IL frameworks (e.g., \cite{gervet2023act3d,xian2023chaineddiffuser,ma2024hierarchical}) by treating learned keyposes as terminal control targets tracked via keypoint-based admittance control.
In contrast, our approach leverages keyposes as generative trajectory guides, which explicitly enhance multi-stage success rates. This functional dichotomy in keypose utilization constitutes the core distinction between our method and Bi-KVIL.

\begin{figure*}[!ht]
    \centering
    \includegraphics[width=0.85\textwidth]{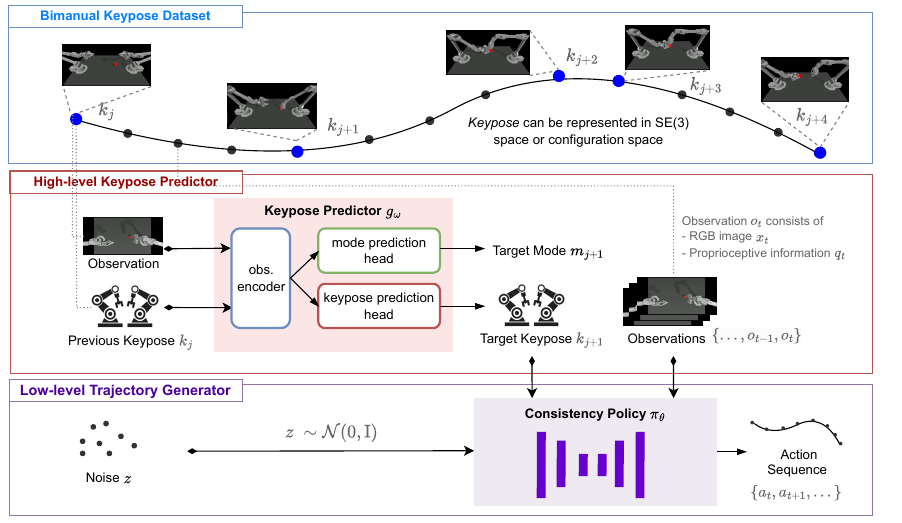}
    \vspace{-2mm}
    \caption{Framework of bimanual keypose-conditioned consistency policy.}
    \label{fig:overview}
    \vspace{-4mm}
\end{figure*}

\section{Preliminaries: Consistency Model} 
\label{sec:preliminary}

In this section, we first briefly introduce CMs as a generative modeling approach to fit data distributions~\cite{song2023consistency,song2024improved}. Then we show that its \textit{one-step generation property} can be leveraged in IL settings as robot visuomotor policies to model observation-conditioned action distributions, compared with their iterative counterparts, DDPMs~\cite{dhariwal2021diffusion, song2020scorebased}.

\paragraph{Modeling Distribution with Consistency Models}
As a score-based generative modeling method, CMs are proposed to simplify the iterative sampling in DDPMs to accelerate generation. 
CMs are built on the fact that with a special noise schedule, the diffusion process turns into a probability-flow ordinary differential equation:
$\dif\rvx = \left[ -\sigma\nabla \log p_\sigma(\rvx) \right]\dif\sigma$, where $\sigma \in [\epsilon, \sigma_\text{max}]$ and $p_\sigma(\rvx)$ is the distribution of noised sample $\rvx^\sigma$ at a continuous noise level $\sigma$; the score function $\nabla\log{p_\sigma(\rvx)}$ is the gradient of $p_\sigma(\rvx) = \int \pdata(\rvy)\gN(\rvx|\rvy,\sigma^2 \mI)\dif\rvy$; $\epsilon$ is a fixed small positive number such that $p_{\epsilon}(\rvx) \approx p_{\text{data}}(\rvx)$ and $\sigma_\text{max}$ is sufficiently large such that $p_{\sigma_\text{max}}(\rvx) \approx \gN(0,\sigma_\text{max}^2 \mI)$. 
The pros of an ODE is that the original data $\rvx^\epsilon$ can be retrieved given any $(\rvx^\sigma, \sigma)$ pair, by reversing the ODE trajectory. 
In other words, there is an operation $\vf$ mapping all diffused samples along the same trajectory back to their noiseless origin $\rvx$, which is called \textit{self-consistency condition} in~\cite{song2023consistency}:
\begin{equation}
\begin{aligned}
    \vf(\rvx^{\sigma_\text{max}}, \sigma_\text{max}) &= \vf(\rvx^{\sigma}, \sigma) = \cdots \\
    &= \vf(\rvx^{\epsilon}, \epsilon) = \rvx^\epsilon, \quad \forall \sigma \in [\epsilon, \sigma_\text{max}].
    \label{eq:self_consistency}
\end{aligned}
\end{equation}

\paragraph{Training and Inference of Consistency Models} 
A parameterized approximated CM, typically represented by neural network $\vf_{\bm{\theta}}$, can be trained by enforcing Equation \ref{eq:self_consistency}. 
Discretizing the noise trajectory into $\eps=\sigma_0<\cdots<\sigma_n<\cdots<\sigma_N=\sigma_\text{max}$, we can draw a sample at the noise level $\sigma_n$ by computing $\rvx+\sigma_n\rvz$ where $\rvz \sim \gN(\bm{0},\mI)$. 
Any sample-noise pair on the same ODE trajectory leads to the origin $\rvx^\epsilon$, implying that the outputs of two adjacent noised samples should be identical.
Accordingly, we approximately enforce Equation \ref{eq:self_consistency} by minimizing their differences. Thus, the training objective of CMs is denoted as 
\begin{equation}
\begin{aligned}
\label{eq:cm_loss}
\min_{\bm{\theta}} \gL({\bm{\theta}}, {\bm{\theta}}^{-}) = \E \lbrack \lambda(\sigma_n) d(
&\vf_{\bm{\theta}}(\rvx+\sigma_{n+1}\rvz, \sigma_{n+1}),\\
&\vf_{\bm{\theta}^-}(\rvx + \sigma_n \rvz, \sigma_n)) \rbrack, \notag
\end{aligned}
\end{equation}
where $\lambda$ is a weighting function, $d$ is a distance metric, $n$ is a random integer between $1$ and $N-1$ and $\bm{\theta}^-$ is the exponential moving average (EMA) of $\bm{\theta}$ for training stability. 
With a well-trained CM $\vf_{\bm{\theta}^*}$, we can generate samples by drawing from a Gaussian $\hat{\rvx}^{\sigma_\text{max}}\sim\gN(\bm{0},{\sigma_\text{max}}^2\mI)$ and performing a one-step inference $\hat{\rvx}^\eps=\vf_{\bm{\theta}}(\hat{\rvx}^{\sigma_\text{max}},{\sigma_\text{max}})$. Further details regarding CM design can be found in Appendix~\ref{sec:cm_details}.

\paragraph{Consistency Models as Visuomotor Policies} 
Analogous to Diffusion Policy~\cite{chi2023diffusion}, CMs can be adapted to IL, serving as robot visuomotor policies~\cite{prasad2024consistency}. 
In this case, a CM approximates the conditional distribution $p(\rva_t|\rvo_t)$ of action sequences $\rva_t$ given observations $\rvo_t$ in demonstrations.
At the $t^{\text{th}}$ step of closed-loop control, a consistency policy $\bm{\pi_{\theta}}$ predicts an action sequence by evaluating $\rva_t=\bm{\pi}_{\bm{\theta}}(\hat{\rva_t}^{\sigma_\text{max}},{\sigma_\text{max}}\mid \rvo_t)$ once. 
The key distinguishing feature of consistency policies, compared to diffusion policies, is their ability to generate an action sequence through only one-step inference without iterative sampling, thereby being faster while maintaining modeling quality.
This makes consistency policies computationally efficient during execution, enabling them to complete dynamic tasks besides quasi-static ones.

\begin{figure*}[!ht]
    \centering
    \includegraphics[width=0.88\textwidth]{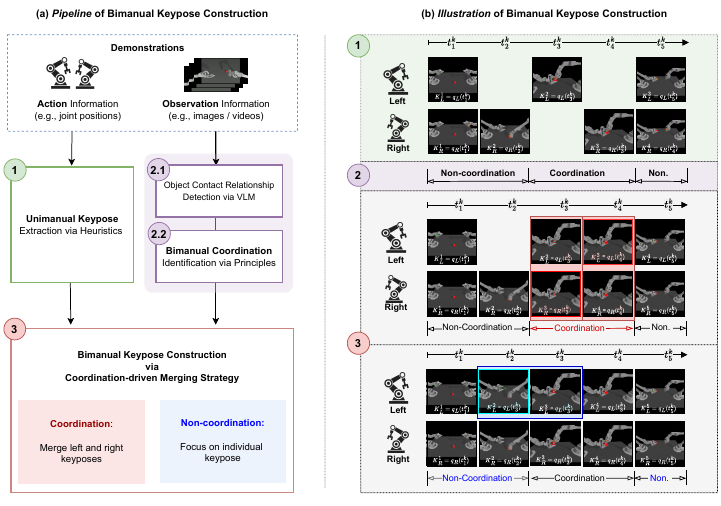}
    \vspace{-3mm}
    \caption{Three-stage pipeline for constructing bimanual keypose dataset.}
    \label{fig:kp_pipeline}
    \vspace{-4mm}
\end{figure*}

\section{Bimanual Keypose-Conditioned Coordination-Aware Consistency Policy} 
\label{sec:method}

This section depicts BiKC+, a hierarchical imitation learning framework, composed of a keypose predictor for high-level guidance and a trajectory generator for low-level behavior, as illustrated in Figure~\ref{fig:overview}. 
We begin with an overview of the hierarchical framework along with notation definitions. Subsequently, we delve into the specifics of each component, detailing the pipeline of bimanual keypose identification, followed by the design of the keypose predictor and the trajectory generator. 
Finally, we conclude this section by outlining the implementation details.

\subsection{Framework Overview and Notation Description} 

We assume a set of demonstrated trajectories $\gD = \{\tau^i \}_{i=0}^{D-1}$, where $\tau^i$ is the $i$-th sequence of observation-action pairs, i.e., $\tau^i = \{ (\vo_t^i, \va_t^i) \}_{t=0}^{T_i-1}$. 
Here, the observation $\vo_t^i = (\vx_t^i, \vq_t^i)$ includes RGB images $\vx_t^i$ and proprioceptive information $\vq_t^i$ (e.g. joint positions), while the action $\va_t^i$ is the desired proprioceptive status. 
Note that we omit superscript $i$ for simplicity if the context is clearly within one trajectory.

For each trajectory $\tau$, we identify a sequence of bimanual keyposes denoted as {$\{\vk_0, \cdots, \vk_j, \cdots, \vk_{J-1}\}$}. 
Each keypose $\vk_j$ is assigned a bimanual coordination indicator $\evm_j$, where {$\evm_j=1$} indicates that the bimanual keypose requires temporal and spatial synchronization, and {$\evm_j=0$} indicates otherwise.
These keyposes refer to shared commonalities across different trajectories of a specific bimanual manipulation task.
They generally indicate the completion of a subtask~\cite{garrett2021integrated} and establish the pre-condition for the subsequent subtask, thereby serving as sub-goals within a long-term task.
This is illustrated by a \textit{Transfer} task shown at the top of Figure~\ref{fig:overview}, where keyposes include grasping the cube with the right arm and then passing it to the left arm.
Keypose can be represented in either SE(3) or configuration space. We choose the latter in this work (i.e., joint positions of bimanual arms) to align with the representation of states and actions.

Based on the demonstration set and the keypose set, we can learn a high-level keypose predictor $\vg_{\bm{\omega}}(\hat{\vk}, \hat{\evm} \mid \vo, \vk)$ and a low-level trajectory generator $\bm{\pi}_{\bm{\theta}}( \rva \mid \rvo, \hat{\vk})$. 
As depicted in Figure~\ref{fig:overview}, the keypose predictor $\vg_{\bm{\omega}}(\cdot, \cdot \mid \vo, \vk)$ is responsible for updating the target keypose $\hat{\vk}$ and its corresponding indicator $\hat{\evm}$ according to the current observation and the previously established keypose.
Notably, this update triggers only upon successful attainment of the current target keypose.
Under the guidance of the target keypose, the trajectory generator enhances its awareness of critical steps required to complete each sub-stage task. It generates a short-horizon action sequence conditioned on both the updated target keypose and historical observations.

\begin{figure}[!t]
    \includegraphics[width=0.88\linewidth]{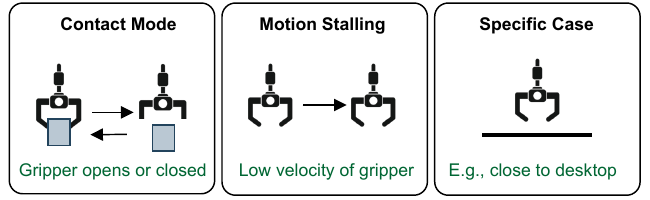}
    \vspace{-2mm}
    \caption{Heuristics rules for identifying unimanual keypose.}
    \label{fig:kp_stage1}
    \vspace{-4mm}
\end{figure}

\subsection{Bimanual Keypose Identification} 

While keyposes have been established for unimanual manipulation \cite{gervet2023act3d,xian2023chaineddiffuser,ma2024hierarchical}, their extension to bimanual coordination scenarios remains underexplored. To address this gap, we introduce a novel bimanual keypose identification method featuring automated coordination mode detection as shown in Figure \ref{fig:kp_pipeline}(a), which includes three stages:
 \begin{enumerate}
    \item Unimanual keypose extraction through heuristic analysis of action-space demonstration data (e.g., joint positions);
    \item Bimanual coordination mode identification via physical contact relations among task objects and robotic arms, using state-space demonstration data (e.g., RGB observations);
    \item Bimanual keypose dataset construction through novel coordination-driven merging strategies.
\end{enumerate}
Our key insight is to integrate an understanding of bimanual coordination into the construction of keypose datasets. This approach considers both robot-centric action features and task-centric operation styles, leading to comprehensive yet flexible keypose identification for cooperative dual-arm movements. The details of each stage are as follows:

\paragraph{\textbf{Stage-1. Unimanual Keypose Extraction}} 

We begin by separately extracting keyposes for each arm with heuristic rules.
Several heuristics have been summarized from previous studies \cite{gervet2023act3d, xian2023chaineddiffuser, ma2024hierarchical} to determine whether an arm is at a keypose. As shown in Figure~\ref{fig:kp_stage1}, these heuristic rules include: (1) changes in robot-object contact modes (i.e., grasping or releasing); (2) stalling of motion before interaction; 
and (3) case-specific rules based on spatial-temporal relationships between arms and/or objects, such as the distance between two grippers or their respective heights relative to the tabletop.
To implement this, we scan the action trajectory in each demonstration to assess whether the following criteria are met: (1) the gripper either begins to open or stops closing; (2) the gripper's velocity drops below a specified threshold; and (3) the relative distance or height is less than a predetermined threshold. Timesteps that satisfy these conditions are labeled as keypose steps.

Directly transferring heuristics from unimanual to bimanual manipulation is challenging because a bimanual system must coordinate both arms to perform shared tasks. During unimanual tasks, one arm can immediately proceed to the next keypose upon achieving the current one. In bimanual tasks, however, the arms must interact and maintain synchronization at keyposes. While prior research proposes a comprehensive taxonomy of bimanual coordination \cite{krebs2022bimanual} to determine whether an arm should pause or move at a specific keypose, applying this taxonomy for keypose identification and prediction requires significant manual hard-coding and annotation. Consequently, this approach becomes less systematic and difficult to generalize. To address these limitations, we propose a solution that automatically identifies the bimanual coordination mode based on object-centric contact relationships, as detailed in the following section.

\begin{figure*}[!t]
    \centering
    \includegraphics[width=0.92\textwidth]{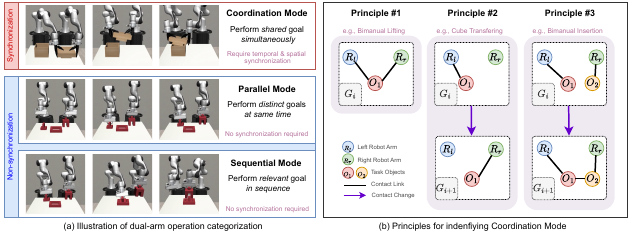}
    \vspace{-2mm}
    \caption{Illustration of dual-arm operation categorization and coordination-mode identification principles}
    \label{fig:kp_stage2}
    \vspace{-4mm}
\end{figure*}

\paragraph{\textbf{Stage-2. Coordination Mode Identification}}

We first categorize bimanual operations and analyze their relationship to bimanual keypose construction, then detail our solution for automatically identifying coordination modes.

Following \cite{jiang2024dexmimicen}, we classify bimanual operations into three categories (Figure \ref{fig:kp_stage2}(a)):
\begin{itemize}
    \item \textit{Coordination Mode}: Both arms operate simultaneously toward a shared goal that requires spatial and temporal synchronization (e.g., jointly lifting an object where unsynchronized movement causes failure).
    \item \textit{Parallel Mode}: Each arm pursues distinct goals concurrently without synchronization requirements (e.g., simultaneously grasping separate objects independently).
    \item \textit{Sequential Mode}: Arms complete relevant goals in a specific order without synchronization (e.g., assembling components where actions must sequence but may not synchronize, such as ordered workpiece insertion).
\end{itemize}
Only the coordination mode demands strict spatio-temporal synchronization. The other two modes allow asynchrony in either temporal or spatial domains.

We propose a VLM-assisted contact-aware solution to automatically identify bimanual coordination modes.
First, we use a pre-trained vision-language model (VLM), specifically Qwen2.5-VL-72B-instruct, to detect changes in object contact relationships. The VLM takes a sequence of image frames from demonstration data as input and outputs a sequence of contact graphs ${G_i}$, where edges represent contacts.
Second, using these extracted contact relationships, we identify the bimanual coordination mode by applying three principles, illustrated in Figure \ref{fig:kp_stage2}(b). The first principle detects coordination using static contact states, while the other two identify coordination through contact-relationship changes:
\begin{itemize}
    \item Principle \textbf{\#1}: 
    When the left robot arm $R_l$ and the right robot arm $R_r$ make contact with the same object $O_1$ simultaneously, these two arms are in a coordination mode (e.g., lifting a box).
    \item Principle \textbf{\#2}: 
    In the $i^{\text{th}}$ graph $G_i$, object $O_1$ is in contact with $R_l$; in the subsequent graph $G_{i+1}$, $O_1$ loses contact with $R_l$ and establishes new contact with $R_r$. This shift in object contact relationships indicates a bimanual coordination mode (e.g., transferring a cube).
    \item Principle \textbf{\#3}: 
    In the $i^{\text{th}}$ graph ${G_i}$, $O_1$ contacts $R_l$ and $O_2$ contacts $R_r$; in the next graph $G_{i+1}$, $O_1$ and $O_2$ form a new contact. Here, the direct contacts between the arms and objects remain unchanged, but the emergence of contact between $O_1$ and $O_2$ integrates the two arms into a coordinated mode (e.g., two-arm insertion).
\end{itemize} 
Here, $R_l$ and $R_r$ represent the left and right robot arms, respectively, and $O_i$ denotes manipulated objects. Durations satisfying these principles define the coordination mode. Keyposes within this mode must be synchronized to ensure precise alignment of both arms during coordinated object manipulation.

These three coordination identification principles cover the robotic tasks studied in this paper, while remaining inherently extensible to novel scenarios. This adaptability strengthens our solution's robustness and broadens its applicability across diverse robotic tasks.

\paragraph{\textbf{Stage-3. Coordination-Driven Merging }}

We propose a Coordination-driven Merging strategy to construct bimanual keypose datasets, leveraging the extracted unimanual keyposes and the identified coordination modes. The strategy operates differently across \textit{coordination} and \textit{non-coordination} modes:

\begin{enumerate}
    \item \textit{Merging Dual-Arm Keyposes in Coordination Mode}: 
    When a unimanual keypose occurs during coordination mode, we assume the other arm simultaneously achieves a keypose.
    For example, in Figure \ref{fig:kp_pipeline}(b), the left arm's second keypose $\vk_L^2$ at timestep $t^k_3$ occurs during coordination phase, but no right-arm keypose exists.
    Our strategy then designates the right arm's joint positions
    $q_R(t^k_3)$ as a keypose. This enforced synchronization ensures precise spatiotemporal alignment for shared objectives.
    \item \textit{Preserving Keypose Independence in Non-Coordination Mode}:
    When a keypose occurs during non-coordination mode, the other arm operates independently. As shown in Figure \ref{fig:kp_pipeline}(b), the right arm's keypose $\vk^2_R$ at timestep $t^k_2$ has no corresponding left-arm keypose. Without synchronization requirements, the left arm proceeds freely toward its subsequent keypose at $t^k_3$ maintaining execution flexibility.
\end{enumerate}

This three-step approach successfully identifies bimanual keyposes for all task demonstrations, inherently integrating interaction and synchronization understanding into the bimanual keypose dataset construction.

\begin{algorithm*}[!t] \footnotesize
\centering
\caption{\small{Policy Deployment Workflow}}
\begin{minipage}{\textwidth}
\begin{algorithmic}[1]
    \Require: keypose predictor $\bm{g_{\omega}(o, \vk_{\text{cur}})}$
    \Require: consistency policy $\bm{\pi_{\theta}(o, \vk_{\text{nxt}})}$
    \For{t = 1,2,..., T}
        \If {t==0}
            \State $\bm{o_0} \leftarrow \text{env.reset()}$
            \State $\bm{k_{\text{nxt}}}, \bm{m_{\text{nxt}}} \leftarrow \bm{g_{\omega}(o_0, k_0)}$ \Comment{predict keypose and mode}
        \EndIf
        \State $\bm{a_t} \leftarrow \bm{\pi_{\theta}(o_t,  \vk_{\text{nxt}}})$ \Comment{generate actions conditioned on keypose}
        \State $\Phi_1$ $\leftarrow$ $ \Bigl\{ {\bm{m_{\text{nxt}}}}=1$ \text{and} $\big\|\bm{a_t}[q_\text{L}, q_\text{R}] - \bm{\vk_\text{nxt}}[q_\text{L}, q_\text{R}] \big\| < \varepsilon \Bigl\} $ \Comment{reach both keyposes in coordination mode}
        \State $\Phi_2$ $\leftarrow$ $ \Bigl\{ {\bm{m_{\text{nxt}}}}=0$ \text{and} $\bigl( \big\| \bm{a_t}[q_\text{L}] - \bm{\vk_{\text{nxt}}}[q_\text{L}] \big\| < \varepsilon$  \text{or} $ \big\| \bm{a_t}[q_\text{R}] - \bm{\vk_{\text{nxt}}}[ q_\text{R}] \big\| < \varepsilon \bigl) \Bigl\}$ \Comment{reach either non-coordination keypose}
        \If {$\Phi_1$ \text{or} $\Phi_2$}: 
            \State $\bm{\vk_{\text{cur}}} \leftarrow \bm{k_{\text{nxt}}}$
            \State $\bm{\vk_{\text{nxt}}, m_{\text{nxt}}} \leftarrow \bm{g_{\omega}(o, \vk_{\text{cur}})}$  \Comment{update keypose and mode}
        \EndIf
        \State $\bm{o_{t+1}} = \text{env.step}(\bm{a_t})$
    \EndFor
\end{algorithmic}
\end{minipage}
\label{algo:inference}
\end{algorithm*}

\subsection{Bimanual Keypose Predictor} 

Through the bimanual keypose identification pipeline, we construct a bimanual keypose set $\gK^i=\{ \vk_j^i, m_j^i\}_{j=0}^{J_i}$ for each demonstration $\tau^i= \{ (\vo_t^i, \va_t^i) \}_{t=0}^{T_i-1}$ in set $\gD = \{\tau^i\}_{i=0}^{D-1}$. 
Here $\vk_j^i$ is the $j$-th keypose in the $i$-th trajectory, represented by joint positions such that $\vk_j^i=\vq_{t_j}^i$. And $m_j^i$ is the indicator of bimanual coordination mode corresponding to the keypose  $\vk_j^i$. Note that $t_j$ is the corresponding timestep ($t_0=0, t_{J_i}=T_i$). 

Since the number of keyposes is much smaller than total steps (i.e., $J_i\ll T_i$), we enrich the dataset to enable the keypose predictor to forecast target keyposes at every step, rather than only at the identified keypose steps.
This strengthens the predictor's capability by leveraging a greater amount of data, thereby improving accuracy and reliability.
As a result, the keypose dataset is formulated as
\begin{equation}
    \label{eq:keypose_dataset}
        \gT_{\text{keypose}} = \bigcup\limits_{i=0}^{D-1} \bigcup\limits_{j=0}^{J_i-1} \bigcup\limits_{t=t_j}^{t_{j+1}-1} \left\{ (\vo^i_t, \vk^i_{j}, \vk^i_{j+1}, m^i_{j+1}) \right\} , 
\end{equation}
which consists of tuples including the observation $\vo_t^i$ along with its preceding and succeeding keyposes.

The keypose predictor $\{ \hat{\vk}_{j+1}, \hat{m}_{j+1} \} = \vg_{\bm{\omega}}(\vo_t, \vk_j)$ employs a dual-head neural network. A shared observation encoder branches into two components: a bimanual keypose head designed as a CM~\cite{song2023consistency,song2024improved}, and an MLP-based mode head for binary classification. CM’s ability to model multimodal distributions allows the keypose head to capture diverse potential keyposes from varying demonstrations, preserving robotic flexibility during low-level action generation.
During deployment, predicted keyposes and coordination indicators synergistically guide the low-level policy. If the mode is identified as \textit{coordination}, sub-tasks complete only when both arms simultaneously reach their keyposes. If the mode is \textit{non-coordination}, sub-tasks complete upon either arm reaching its keypose, enabling independent progression.

The training of the keypose head $\hat{\vk}_{j+1}=\vg_{\bm{\omega}}^{\text{k}}(\vo_t, \vk_j)$ is achieved by minimizing a specific consistency training objective $\gL_{\text{k}}$ as follows:
\begin{equation}
    \begin{aligned}
        \gL_{\text{k}}(\bm{\omega}, \bm{\omega}^{-}) = & \mathop{\E}\limits_{(\vo, \vk, \vk') \sim \gT_{\text{keypose}}} \lbrack \lambda(\sigma_n) \cdot d(\vg_{\bm{\omega}}^{\text{k}} (\vk'+ \\
        & \sigma_{n+1}\rvz, \sigma_{n+1} | \rvo, \vk), \vg_{{\omega}^{-}}^{\text{k}} (\vk' + \sigma_n \rvz, \sigma_n |\rvo, \vk) ) \rbrack, \notag
    \end{aligned}
\end{equation}
where $\bm{\omega}^-\gets \textit{stopgrad}(\mu\bm{\omega^-}+(1-\mu)\bm{\omega})$ with EMA decay rate $\mu$ and $\textit{stopgrad}$ denotes the stopping gradient calculation. $\rvz \sim \gN(\vzero, \rmI)$ represents a random variable sampled from a standard normal distribution.
The mode head $\hat{m}_{j+1}=\vg_{\bm{\omega}}^{\text{m}}(\vo_t, \vk_j)$ is trained by minimizing the Binary Cross-Entropy (BCE) loss $\gL_{\text{m}}$ between the predicted mode $\hat{m}$ and the ground truth $m$:
\begin{equation}
    \label{eq:keypose_loss}
    \begin{aligned}
        &\gL_{\text{m}}(\omega) = \mathop{\E}\limits_{m \sim \gT_{\text{keypose}}} \lbrack  -m \cdot \log{\sigma(\hat{m})} - (1-m) \log{(1-\sigma(\hat{m}))} \rbrack, \notag
    \end{aligned}
\end{equation}
where $\sigma(\cdot)$ is a sigmoid function. The keypose predictor is then learned by minimizing the combined loss: 
\begin{equation}
    \gL_{\text{keypose}}(\omega) = \gL_{\text{k}}(\bm{\omega}, \bm{\omega}^{-}) + \beta \gL_{\text{m}}(\omega) \notag
\end{equation}
where $\beta$ is a weighting parameter balancing the contributions of the keypose-head loss and the mode-head loss, with a default value of 1.

\subsection{Keypose-Guided Trajectory Generator} 

We formulate the trajectory generator which plans short-horizon action sequences as a CM~\cite{song2023consistency, song2024improved}, leveraging CM's capability of modeling complex distributions with one-step sampling. In this part, we describe the training of this CM-based policy that models action distribution conditioned on the target keypose and historical observations.

\begin{figure}[t!]
    \centering
    \includegraphics[width=0.95\linewidth]{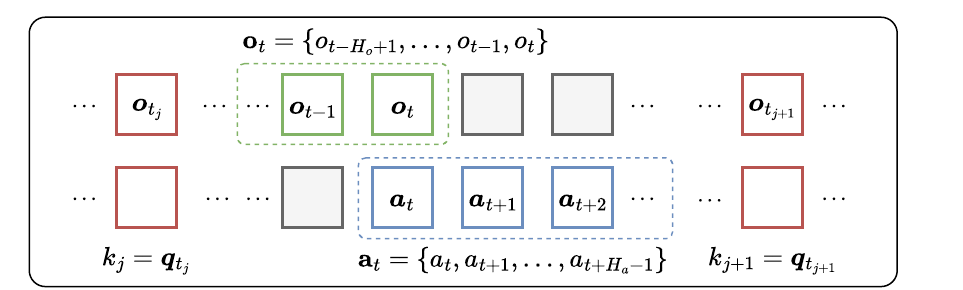}
    \vspace{-1.5mm}
    \caption{Keyposes and trajectories sampled from demonstrations, for training keypose predictor and trajectory generator.}
    \label{fig:traj_sample}
    \vspace{-4mm}
\end{figure}

The keypose-conditioned trajectory generator $\bm{\pi}_{\bm{\theta}}$ is trained on an action sequence dataset built on the demonstrations $\gD = \{\tau^i\}_{i=0}^{D-1}$ and the keyposes $\{ \gK^i \}_{i=0}^{D-1}$, as illustrated in Figure~\ref{fig:traj_sample}. The dataset can be represented as
\begin{equation}
    \label{eq:traj_dataset}
    \gT_{\text{traj}} = \bigcup\limits_{i=0}^{D-1} \bigcup\limits_{j=0}^{J_i-1} \bigcup\limits_{t=t_j}^{t_{j+1}-1} \left\{ ( \rvo^i_t, \vk^i_{j+1}, \rva^i_t) \right\}, \notag
\end{equation}
where $\rvo_t = \{ \vo_{t-H_{o}+1}, \dots, \vo_{t-1}, \vo_t \}$ and $\rva_t = \{\va_t, \va_{t+1}, \dots, \va_{t+H_{a}-1} \}$. Here, $H_o$ and $H_a$ denote the length of historical observations and action sequence, respectively. The target keypose $\vk^i_{j+1} = \vq^i_{t_{j+1}}$ is the joint positions at the next keypose step exactly after $t$.
A special case is that if the indices of actions exceed the index of the target keypose (that is, $t+H_a-1 \ge t_{j+1}$), we pad actions after $t_{j+1}$ with $\va_{t_{j+1}-1}$ to make the two arms staying at the keypose. Therefore, all short-horizon trajectories to be imitated will not cross sub-stage boundaries.

The keypose-conditioned trajectory generator $\bm{\pi}_{\bm{\theta}}( \rvo_t, \vk_{j+1})$ is learned by minimizing the consistency training objective
\begin{equation}
    \begin{aligned}
    \label{eq:traj_loss}
        &\gL_{\text{traj}}(\bm{\theta}, \bm{\theta^{-}}) = \mathop{\E}\limits_{\gT_{\text{traj}},n} \lbrack \lambda(\sigma_n) \cdot \\
        &d(\bm{\pi}_{\bm{\theta}}(\rva+ \sigma_{n+1}\rvz, \sigma_{n+1} | \rvo, \vk), \bm{\pi}_{\theta^{-}}(\rva + \sigma_n \rvz, \sigma_n |\rvo, \vk)) \rbrack, \notag
    \end{aligned} 
\end{equation}
where $\bm{\theta}^-\gets \textit{stopgrad}(\mu\bm{\theta^-}+(1-\mu)\bm{\theta})$ with EMA decay rate denoted by $\mu$ and $\textit{stopgrad}$ meaning stopping gradient calculation; $\rvz \sim \gN(\vzero, \rmI)$. 
During training, the keypose $\vk$ fed into $\bm{\pi}_{\bm{\theta}}$ is the extracted ground truth; while during policy deployment, it is predicted by the predictor $\bm{g}_{\bm{w}}$. 
The workflow for policy deployment, which utilizes predicted keyposes to guide action generation and the predicted mode as a bimanual coordination indicator, is detailed in Algorithm \ref{algo:inference}.

\subsection{Implementation Details} 
We train the keypose predictor $\bm{g}_{\bm{w}}$ and the trajectory generator $\bm{\pi}_{\bm{\theta}}$ separately and chain them during deployment. 
For the keypose predictor, each $640\times480$ RGB image is encoded by a non-pretrained ResNet-18 encoder~\cite{he2016deep} into a feature sequence. 
When multiple cameras are involved, the feature sequences from each camera are concatenated along the length dimension, resulting in a comprehensive vision feature sequence.
Both $\bm{g}_{\bm{w}}$ and $\bm{\pi}_{\bm{\theta}}$ leverage 1D UNet architectures similar to that described in~\cite{chi2023diffusion}. 
To train the CM-based models, we use several improved techniques proposed in \cite{song2024improved} to enhance the policy performance. 
Specifically, the weighting function is $\lambda(\sigma_n) = 1/(\sigma_{n+1}-\sigma_n)$ while the distance metric is Pseudo-Huber loss $d(\vx,\vy)=\sqrt{\Vert \vx-\vy\Vert^2_2+c^2}-c$ with $c=0.0064$. Furthermore, we adopt $\mu=0$ and a step-shaped curriculum for discretization steps $N$.
Additionally, for both keypose predictor and consistency policy, each input element is uniformly normalized into $[-1,1]$ while the output is denormalized back to the original ranges. 
Both models are trained using AdamW~\cite{loshchilov2018decoupled} optimizer, with a batch size of $B=64$ and a learning rate cosine annealing from 1e-4 to 0. More detailed hyper-parameters are provided in Appendix~\ref{sec:hyper_params}.

During policy deployment (Algorithm \ref{algo:inference}), the threshold $\varepsilon$ determines successful keypose attainment. This critical value was empirically set to $0.06$ for simulation and $0.2$ for real-world experiments to ensure reliable policy execution.

\begin{figure*}[!t]
    \centering
    \includegraphics[width=0.77\textwidth]{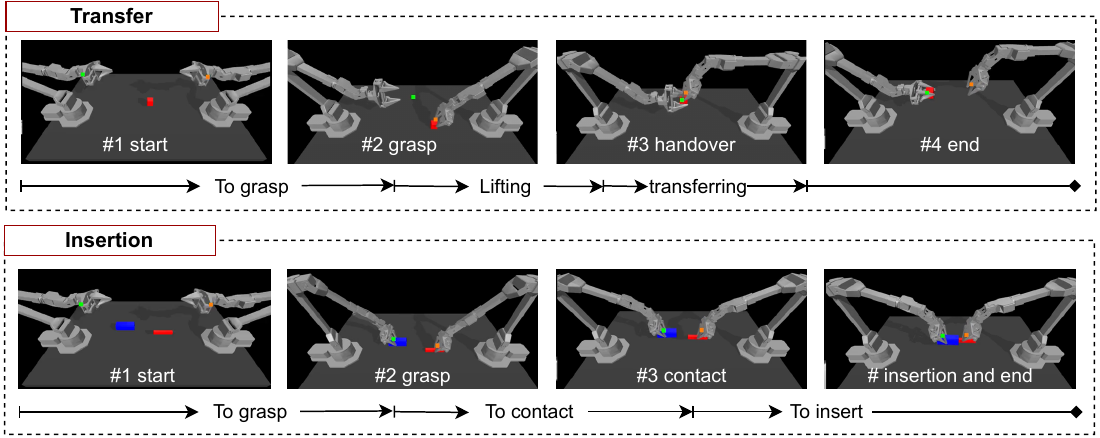}
    \vspace{-1mm}
    \caption{Illustration of simulated bimanual multi-stage tasks (Transfer and Insertion), with predicted keyposes highlighted.}
    \vspace{-3mm}
    \label{fig:sim_results}
\end{figure*}

\section{Experiments} 
\label{sec:experiments}

We evaluate BiKC+ on both simulated and real-world manipulation tasks to address the following research questions:
1) Does the integration of keyposes enhance the performance in bimanual multi-stage manipulation tasks?
2) Are CM-based policies effective in reducing inference latency?
3) Can BiKC+ increase the overall success rate and operational efficiency in real-world multi-stage tasks?
We compare BiKC+ with state-of-the-art (SOTA) imitation learning baselines, namely ACT~\cite{zhao2023learning} and DP~\cite{chi2023diffusion}, previously detailed in Section \ref{sec:related}. In addition, CP (Consistency Policy) serves as a non-keypose ablation. It employs a consistency model \cite{song2023consistency, song2024improved}, trained from scratch using the consistency loss as defined in~\eqref{eq:cm_loss}.

\subsection{Simulated Experiments} 

\subsubsection{Tasks and Settings}
We evaluate BiKC+ on two tasks built in MuJoCo~\cite{todorov2012mujoco}, specifically \textit{Transfer} and \textit{Insertion} developed by~\cite{zhao2023learning}. Both tasks involve multiple stages, requiring precise coordination between two arms, as illustrated in Figure~\ref{fig:sim_results}.
In the simulation environment, two Interbotix vx300s arms are positioned on opposite sides of a tabletop. The observation $\vo_t$ includes a $480 \times 640$ RGB image captured from a static top-down camera and a 14D vector recording the joint positions of both arms. The action is also a 14D vector, indicating the target joint positions in the next timestep.
During evaluation, the initial positions of the cube, socket, and peg are randomly sampled from a predefined rectangular area. In this experiment, we utilized 50 demonstrations provided by Zhao et al.~\cite{zhao2023learning} to train all models across each task. These demonstrations were collected by a scripted policy at a frequency of 50Hz, lasting 400 timesteps, equivalent to 8 seconds. In simulation, our main objective is to examine the impact of keyposes on success rates. It is important to note that we do not assess inference latency here, as the simulation is paused during policy inference.

\begin{table}[t!]
\centering
\caption{Overall success rate (\%) of each sub-stage in two simulated tasks, with top-two results highlighted.}
\label{tab:sim_results}
\begin{tabular}{clccc|c}
\toprule
\textbf{Task}             & \textbf{Sub-stage} & \textbf{ACT} & \textbf{ DP} & \textbf{CP} & \textbf{BiKC+} (ours) \\
\midrule
\multirow{3}{*}{Transfer} & To grasp           & \textbf{97}  & 96   & 96 & \textbf{98} \\
                          & Lifting            & 90           & \textbf{96}       & 93 & \textbf{98} \\
                          & Transferring       & 86           & \textbf{96}       & 93 & \textbf{98} \\
\midrule
\multirow{3}{*}{Insertion}& To grasp           & \textbf{93}     & 83             & 83 & \textbf{98} \\
                          & To contact         & \textbf{90}     & 47             & 53 & \textbf{96}  \\
                          & To insert          & 32              & 37             & \textbf{38} & \textbf{43}\\
\bottomrule
\end{tabular}
\vspace{-5mm}
\end{table}

\subsubsection{Results}

Table~\ref{tab:sim_results} presents average success rates for all sub-stages and overall tasks, with results averaged over 3 seeds and 50 rollouts (top two highlighted). 
In both simulated experiments, BiKC+ achieved the highest success rate in each sub-stage task and the overall task, outperforming baseline methods ACT \cite{zhao2023learning} and DP \cite{chi2023diffusion}.
BiKC+ significantly surpasses its non-keypose counterpart CP by 2\%, 5\%, and 5\% in the three sub-tasks of \textit{Transfer}, and by 16\%, 43\%, and 5\% in those of \textit{Insertion}. This improvement is attributed to the guidance provided by keyposes as illustrated. 

\begin{figure}[t!]
    \centering
    \includegraphics[width=0.9\linewidth]{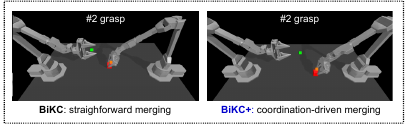}
    \vspace{-3mm}
    \caption{Comparison of bimanual keypose predicted by BiKC \cite{yu2024bikc} and BiKC+.}
    \label{fig:sim_kp}
    \vspace{-4mm}
\end{figure}

We visualize the predicted keyposes in Figure~\ref{fig:sim_results}, segmenting the task into distinct sub-tasks. 
Taking \textit{Transfer} as an example, keyposes are delineated at critical junctures, specifically at the start, upon grasping the cube, handover, and in the retracting stage.
These sequential keyposes consistently emerge across different demonstrations, serving as a rough sketch for the success of the task. 
Additionally, Figure~\ref{fig:sim_kp} visualizes the bimanual keyposes predicted by BiKC \cite{yu2024bikc} and BiKC+, respectively. At this non-coordination stage (specifically, \textit{\#2 grasp}), BiKC+ allows the left arm to concentrate on its independent sub-goal; whereas BiKC imposes a keypose on the left arm, potentially restricting its movement flexibility. 
In this \textit{Transfer} task, BiKC+ outperforms BiKC by 3\% in terms of overall success rate, indicating that the coordination-driven merging strategy proposed in BiKC+ facilitates the reasonable keypose identification and positively impacts the policy performance. 
In summary, keyposes play a vital role in guiding action generation and precisely identifying the completion of sub-stages within a task.

\begin{figure*}[!ht]
    \centering
    \includegraphics[width=0.77\textwidth]{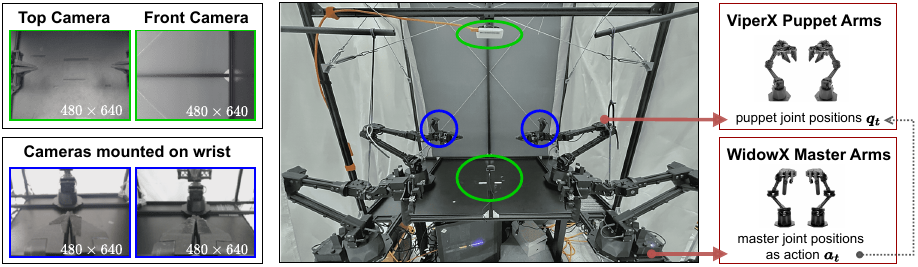}
    \vspace{-3mm}
    \caption{Real-world ALOHA platform.}
    \label{fig:real_platform}
\vspace{-4mm}
\end{figure*}

\subsection{Real-world Experiments} 

We evaluate BiKC+ on four real-world tasks that feature multiple sub-stages, intricate coordination, and rich contact: Screwdriver Packing, Pants Hanging, Placing and Picking on Conveyor, and Cup Insertion.
The objective of \textit{Screwdriver Packing} and \textit{Pants Hanging} is to demonstrate BiKC+'s ability to enhance sub-stage success rate while improving operational efficiency.
\textit{Placing and Picking on Conveyor} involves a dynamic object, which allows us to emphasize the advantage of CM over DP in terms of inference speed.
Finally, we assert that BiKC+ can capture multi-modality in \textit{Cup Insertion}, thereby overcoming the limitations of ACT.

\subsubsection{Platform}
We conducted real-world experiments using a low-cost open-source hardware system for bimanual teleoperation, known as ALOHA~\cite{zhao2023learning}.
ALOHA comprises two 7-DOF master arms, two 7-DOF puppet arms, and various accessories.
During teleoperation for demonstration recording, a human operator manipulates the master arms. The corresponding joint positions are recorded as action $\va_t$ and transmitted to the puppet arms for synchronization and execution.
In the deployment phase, the desired joint positions for the puppet arms are generated by neural policies for autonomous execution. 
Additionally, the system is equipped with four cameras that provide visual observations. Two of these cameras are positioned at the front and top of the frame, offering a static overview of the workspace. The remaining two cameras are mounted on puppet wrists, delivering close-up views essential for fine-grained manipulation.

\begin{figure*}[!ht]
    \centering
    \includegraphics[width=0.83\textwidth]{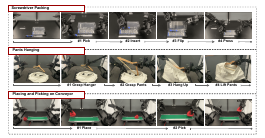}
    \vspace{-3mm}
    \caption{Illustration of three real-world bimanual multi-stage tasks.}
    \label{fig:real_tasks}
    \vspace{-5mm}
\end{figure*}

\subsubsection{Tasks and Settings}

As illustrated in Figure~\ref{fig:real_tasks}, we designed three bimanual multi-stage tasks, each highlighting a specific capability of BiKC+.
\begin{itemize}
    \item \textit{Screwdriver Packing} is a long-horizon and fine-grained task involving several challenging components, such as eye-hand coordination (e.g., picking), high precision (e.g., insertion), bimanual coordination (e.g., flipping and pressing) and transparent object (e.g., box lid). 
    Due to the prolonged duration of this task, errors and delays may accumulate over different stages, in order to assess whether BiKC+ can improve reliability and efficiency.
    \item \textit{Pants Hanging} introduces deformable objects. This task necessitates coordination, where the right arm must pick up and stabilize the hanger before the left arm grasps and drapes the pants over it. The complex dynamics of the pants require precise interaction between the manipulators and the object to prevent undesired deformations, thereby posing significant challenges. The challenges underscore the BiKC+'s capabilities of precision and coordination.
    \item \textit{Placing and Picking on Conveyor} involves handling dynamic objects in two distinct stages. In the first stage, the left arm picks up a bag and places it on a conveyor belt moving at 74 mm/s. As the bag approaches the right arm, the right arm must accurately pick up and lift the bag at the appropriate moment. Delays in decision-making and execution by the right arm, due to neural network inference, can cause missing the bag, leading to failure. This task underscores the advantages of BiKC+ regarding inference speed. 
\end{itemize}

We collected 50 demonstrations for each task, following the experimental setup outlined in previous work [37]. The initial workspace layout is maintained throughout all demonstrations and rollouts.

\subsubsection{Results}

We report all success rates, inference latencies and total durations in Table~\ref{tab:real_results}. 
To quantify the success rate for each sub-stage task, we follow the measurement approach given by~\cite{fu2024mobile}. For each stage, the success rate is calculated as the number of \textit{Success Cases} divided by the number of \textit{Attempts}. Here, \textit{Attempts} denote the number of successful cases from the previous stage, acknowledging that the robot may fail and terminate at any point. By employing this method, we can derive the final success rate for the entire task by taking the product of success rates from all sub-stages.
Additionally, the inference latency refers to the time required for the policy to generate an action sequence, while operational efficiency is measured by the total duration required to complete tasks within a predetermined number of timesteps. 
All reported results represent the average of 20 rollouts, a practice widely adopted in related work \cite{zhao2023learning, chi2023diffusion, xian2023chaineddiffuser, ma2024hierarchical}. The top two results are highlighted in Table~\ref{tab:real_results}.

\begin{figure}[!ht]
    \centering
    \includegraphics[width=\linewidth]{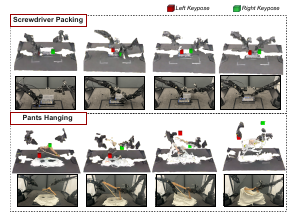}
    \vspace{-5mm}
    \caption{Visualization of predicted keyposes during policy deployment.}
    \label{fig:exp_real_kp}
    \vspace{-5mm}
\end{figure}

In \textit{\textbf{Screwdriver Packing}}, BiKC+ outperforms CP by 44.5\% in terms of success rate, benefiting from guidance by predicted keyposes. The integration of keyposes improves BiKC+'s understanding of the sub-stage goal.
This observation is qualitatively verified in the completion of the final stage (i.e., pressing), as illustrated in Figure~\ref{fig:real_cm_screwdriver}. The CP approach primarily causes the grippers to drift slightly, deviating from the correct maneuver to close the box lid. In contrast, the BiKC+ method successfully accomplishes the task owing to keypose guidance, as illustrated in Figure \ref{fig:exp_real_kp}. 
This outcome highlights the beneficial impact of keypose guidance.
Furthermore, BiKC+ exhibits faster inference than DP, showcasing the advantages associated with CM.
While BiKC+'s success rate and inference latency are similar to those of ACT, its total operation duration (28.0s) is shorter by 9.0s, thereby demonstrating superior efficiency.   
In conclusion, this experiment demonstrates BiKC+'s enhanced performance in reliability and operational efficiency, stemming from its keypose design and CM formulation.

\begin{table}[!t]
\centering
\caption{Success rate (\%), inference latency (ms) and total operation time (s) in real-world tasks. All results are averaged over 20 rollouts.}
\label{tab:real_results}
\begin{tabular}{@{}clccccc@{}}
\toprule
\textbf{Task}              & \textbf{Sub-stage\quad} & \textbf{ACT} & \textbf{DP} & \textbf{CP}  & \textbf{\textit{BiKC+}} \\ 
\midrule
\multirow{7}{*}{\begin{tabular}[c]{@{}c@{}}Screwdriver \\ Packing \end{tabular}} & Pick            &100           & 100     & \multicolumn{1}{c|}{85}       & 85 \\
                          & Insert             & 50           & 30      & \multicolumn{1}{c|}{95}   & 94  \\
                          & Flip               & 100          & 50      & \multicolumn{1}{c|}{90}   & 75  \\
                          & Press              & 100          & 0       & \multicolumn{1}{c|}{18}   & 100 \\ 
                          & \textbf{Overall}     & \textbf{50.0}  & 0.0       & \multicolumn{1}{c|}{15.4}  & \textbf{59.9}    \\ 
                          \cmidrule(lr){2-6}
                          & Infer. Lat.       & 25.6        & 114.3   & \multicolumn{1}{c|}{ 22.6 }        & 34.9 \\
                          & Total Dur.        & 37.0        & 33.8   & \multicolumn{1}{c|}{\textbf{24.8}}  & \textbf{28.0}\\ 
\midrule
\multirow{7}{*}{\begin{tabular}[c]{@{}c@{}} Pants \\ Hanging  \end{tabular}} & Grasp Hanger            & 40          &  80     & \multicolumn{1}{c|}{100}        &     100    \\
                          & Grasp Pant              &    88        &  94       & \multicolumn{1}{c|}{84}    &    90  \\ 
                          & Hang Up              &    14        &  93       & \multicolumn{1}{c|}{81}    &    100  \\ 
                          & Lift Pant           &    100        &  100       & \multicolumn{1}{c|}{100}    &    94.1  \\ 
                          & \textbf{Overall}  &    5.0     &   \textbf{70.0}     & \multicolumn{1}{c|}{65.0}  &    \textbf{85.0}\\
                          \cmidrule(lr){2-6}
                          & Infer. Lat.     &    28.4    &    133.1     & \multicolumn{1}{c|}{21.2}    & 35.5     \\
                          & Total Dur.      &    49.1    &     41.3    & \multicolumn{1}{c|}{\textbf{27.3}}     & \textbf{27.7}  \\
\midrule
\multirow{5}{*}{\begin{tabular}[c]{@{}c@{}} Placing and Picking \\ on Conveyor  \end{tabular}} & Put            & 95          &  100     & \multicolumn{1}{c|}{100}        &     100   \\
                          & Pick              &    0        &  10       & \multicolumn{1}{c|}{100}    &    100  \\ 
                          & \textbf{Overall}  &    0.0     &   10.0     & \multicolumn{1}{c|}{\textbf{100.0}}  &    \textbf{100.0}\\
                          \cmidrule(lr){2-6}
                          & Infer. Lat.     &    27.4    &    115.6     & \multicolumn{1}{c|}{23.3}    & 27.8     \\
                          & Total Dur.      &    29.3    &     25.5    & \multicolumn{1}{c|}{\textbf{18.4}}     & \textbf{19.7}  \\
\bottomrule
\end{tabular}
\vspace{-5mm}
\end{table}

\textbf{
\begin{figure}[!t]
    \centering
    \subfloat[The ``flipping'' and ``pressing'' of Screwdrive Packing]  {\label{fig:real_cm_screwdriver}\includegraphics[width=0.95\columnwidth]{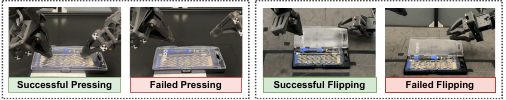}} \\
    \vspace{-0.05in}
    \subfloat[The ``grasp pants'' and ``hang up'' of Pants Hanging]{\label{fig:real_cm_pants}\includegraphics[width=0.95\columnwidth]{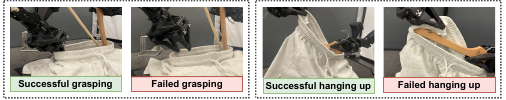}} \\
    \vspace{-0.05in}
    \subfloat[Moments when trying to pick the moving bag on the conveyor]{\label{fig:real_cm_conveyor}\includegraphics[width=0.95\columnwidth]{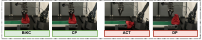}} 
    \caption{Failure examples in real-world experiments.}
    \label{fig:real_cm}
    \vspace{-4mm}
\end{figure}}

\textit{\textbf{Pants Hanging}} is a deformable object manipulation task, which further highlights the critical importance of keypose guidance and the efficiency of CM-based policy.
In each sub-stage task, BiKC+ achieves a success rate exceeding 90\%. In contrast, baseline methods occasionally fail owing to the lack of intermediate keyposes.
The third sub-stage is particularly challenging when the left gripper pulls the pants onto the hanger, primarily due to the garment's unpredictable deformation. For instance, insufficient stretching may cause the pants' edge to snag on the hanger's hook (as depicted in Figure~\ref{fig:real_cm_pants}), whereas excessive stretching might compress the pants' openings. 
As depicted in the third frame of the \textit{Pants Hanging} sequence of Figure \ref{fig:exp_real_kp}, the keypose at this stage  is critical for regulating stretching intensity, thereby enabling BiKC+ to attain superior performance.
Additionally, the comparable performance of the three diffusion-based methods (DP, CP, and BiKC+) in this sub-task underscores their modeling capabilities, presenting an advantage over cVAE used by ACT.
And compared to DP, CM-based policies feature faster inference and improved operational efficiency.

In \textit{\textbf{Placing and Picking on Conveyor}}, both ACT and DP struggle to pick up the bag due to excessive inference latency, indicating their ineffectiveness in manipulating moving objects. 
In contrast, BiKC+ and CP achieve 100\% success rates, attributed to the CM formulation that enables action sequence generation through one-step inference with a latency at 0.01-second level. 
This experiment highlights the advantages of consistency models and emphasizes the critical role of inference speed, as illustrated in Figure~\ref{fig:real_cm_conveyor}. 
It is notable that ACT achieves longer operation durations compared to BiKC+ and CP, despite having similar inference speeds. This is due to ACT's granular prediction process and the utilization of the temporal ensembling (TE) technique. Specifically, ACT predicts an action sequence at \textit{each} timestep, and TE calculates a weighted average over historical predictions, which significantly increases decision delay. In contrast, CP and BiKC+ infer only every $H$ steps, enabling latency amortization and real-time responses.

In summary, BiKC+ shows enhanced success rates and efficiency in bimanual multi-stage tasks, surpassing performances of ACT and DP, as well as the ablated CP.

\begin{table}[!t]
\centering
\caption{Percentage (\%) of behaviour modal in Cup Insertion tasks. All results are averaged over 20 rollouts.}
\label{tab:real_cup_modal}
\begin{tabular}{@{}clccccc@{}}
\toprule
\textbf{Task}                                                                   
& \multicolumn{1}{|c|}{\begin{tabular}[c]{@{}c@{}} \textbf{Behaviour} \\ \textbf{Modal}  \end{tabular}}
&\begin{tabular}[c]{@{}c@{}}   \textbf{BiKC+:} \textit{Coordination}\\ \textit{-driven Merging} \end{tabular}
& \begin{tabular}[c]{@{}c@{}}  \textbf{BiKC:} \textit{Straightforward} \\ \textit{Merging} \end{tabular} \\ 
\midrule
\multirow{3}{*}{\begin{tabular}[c]{@{}c@{}} Cup \\ Insertion  \end{tabular}}    
& \multicolumn{1}{|c|}{Left First}    & 40\%    & 10\% \\
& \multicolumn{1}{|c|}{Right First}   & 55\%    & 90\%  \\
& \multicolumn{1}{|c|}{Together}      & 5\%     & 0\% \\
\bottomrule
\end{tabular}
\end{table}

\begin{figure}[!t]
    \centering
    \includegraphics[width=\linewidth]{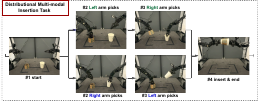}
    \vspace{-0.3in}
    \caption{Cup-insertion task with distributional multi-modal demonstrations.}
    \vspace{-0.2in}
    \label{fig:exp_real_cup}
\end{figure}

\subsection{Further Evaluations and Discussion} 

In this section, we design a \textit{Cup Insertion} task involving diverse and partially conflicting demonstrations to evaluate BiKC+'s capabilities in multi-modality modeling. Additionally, we present BiKC+'s generalization performance across varying objects and under noisy background conditions.

\subsubsection{Multi-modality Modeling Performance}

As illustrated in Figure~\ref{fig:exp_real_cup}, the picking stage presents two distinct demonstration modalities: one where the left arm picks first (25 demonstrations), and another where the right arm picks first (25 demonstrations). 
This necessitates the robot to learn two behavioral modalities conditioned on identical initial observations (as shown in ``\#1 start"), which exemplifies BiKC+'s multi-modality modeling capability.
Table~\ref{tab:real_cup_modal} presents the percentage of each modality across 20 execution rollouts. BiKC+ successfully executed both modalities of behaviors, with percentages of 40\% and 55\% respectively, thereby demonstrating its multi-modality modeling proficiency. Furthermore, occasional instances where both arms grasped objects almost simultaneously further underscore BiKC+'s execution flexibility during non-coordination phase.

Our findings indicate that both the design of the bimanual keypose dataset and the development of its corresponding predictor significantly impact multi-modal and flexible execution performance. 
Specifically, we compared two approaches regarding bimanual keypose construction and keypose predictor design:
\begin{itemize}
    \item \textbf{BiKC+} adopts a coordination-driven merging Strategy for constructing the bimanual keypose dataset coupled with a CM-based keypose predictor;
    \item \textbf{BiKC} employs a straightforward merging mechanism, disregarding the bimanual coordination mode. In this baseline, the keypose predictor is formulated as a Transformer neural network without diffusion-based design.
\end{itemize}
As shown in Table \ref{tab:real_cup_modal}, our design generated both behavioral modalities in nearly equal proportions. In contrast, the baseline exhibited a strong bias towards one specific behavior. This result emphasizes that both the keypose predictor and the trajectory generator must possess multi-modality modeling capabilities. Moreover, our design of keypose datasets preserves execution flexibility, confirming that an understanding of the bimanual coordination mode is crucial for constructing effective bimanual keyposes.

\begin{table}[!t]
\centering
\caption{Success rate (\%) of varying objects in real-world Cup Insertion tasks. All results are averaged over 10 rollouts.}
\label{tab:real_cup_varying_obj}
\begin{tabular}{@{}clccccc@{}}
\toprule
\textbf{Task}              
& \multicolumn{1}{|c|}{\textbf{Sub-stage\quad}}
& \textbf{\begin{tabular}[c]{@{}c@{}} Original \\ Cup  \end{tabular}}   
& \textbf{\begin{tabular}[c]{@{}c@{}} Color-variant \\ Cup  \end{tabular}}   
& \textbf{\begin{tabular}[c]{@{}c@{}} Size-variant \\ Cup  \end{tabular}}   \\ 
\midrule
\multirow{4}{*}{\begin{tabular}[c]{@{}c@{}} Cup \\ Insertion  \end{tabular}}      
                          & \multicolumn{1}{|c|}{Pick A}          &    100   &  90    & 100 \\
                          & \multicolumn{1}{|c|}{Pick B}          &    100   &  89    & 100 \\
                          & \multicolumn{1}{|c|}{Insert}          &   100    &  100   & 80  \\
                          \cmidrule(lr){2-5}
                          & \multicolumn{1}{|c|}{\textbf{Overall}} &   100   &  80  & 80 \\               
\bottomrule
\end{tabular}
\vspace{-2mm}
\end{table}

\begin{figure}[!t]
    \centering
    \includegraphics[width=0.95\linewidth]{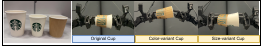}
    \vspace{-3mm}
    \caption{Evaluation of generalization to varying objects in Cup Insertion Task.}
    \label{fig:exp_real_cup_varying_obj}
    \vspace{-4mm}
\end{figure}

\subsubsection{Generalization Performance}

We use \textit{Cup Insertion} as the example task to evaluate BiKC+'s performance in new scenarios, i.e.,  varying object and changing background.

As depicted in Figure \ref{fig:exp_real_cup_varying_obj}, we evaluated BiKC+'s performance with cups that vary in color and size. Table \ref{tab:real_cup_varying_obj} shows that BiKC+ achieved an 80\% success rate across 10 rollouts for both color-variant and size-variant objects. Furthermore, the sub-stage performance statistics in Table~\ref{tab:real_cup_varying_obj} reveal that object color more significantly impacted the ``pick" stage, whereas object size had a greater influence on the ``insert" stage. 
This observation aligns with the distinct features that each sub-stage prioritizes, which are determined by its specific sub-goal.
Hence, while BiKC+ achieved a 100\% success rate with the original object, its generalization performance to varying objects, exceeding 80\%, is considered acceptable.

\begin{figure}[!t]
    \centering
    \includegraphics[width=\linewidth]{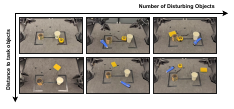}
    \vspace{-0.3in}
    \caption{Illustration of the noisy-background scenarios for Cup Insertion Task.}
    \label{fig:exp_real_cup_noisy_bg}
    \vspace{-4mm}
\end{figure}

As illustrated in Figure \ref{fig:exp_real_cup_noisy_bg}, we evaluated BiKC+'s performance in noisy backgrounds containing disturbing objects. 
We report the success rate for each scenario across 10 rollouts, considering both the number of disturbing objects and their proximity to the task objects. 
Results presented in Table \ref{tab:real_cup_noisy_bg} reveal that the success rate ranges from 60-70\% in distant-proximity scenarios while dropping to 40-50\% in close-proximity scenarios.
We found that the distance between disturbing objects and task objects significantly impacts BiKC+'s performance, and the failure of the picking stage surges when a distracting object gets closer to the object being manipulated. We attribute this to the larger occupation of distracting objects in the wrist camera.
Conversely, within a similar distance range, the number of disturbing objects did not substantially affect BiKC+'s performance.
BiKC+ achieved an average success rate of 56.6\% in noisy background conditions using a zero-shot policy. This performance can be improved to 95\% by incorporating these noisy scenarios into policy training.

\begin{table}[!t]
\centering
\caption{Success rate (\%) for real-world Cup Insertion tasks with disturbing objects. All results are averaged over 10 rollouts.}
\label{tab:real_cup_noisy_bg}
\begin{tabular}{@{}clccccc@{}}
\toprule
{Task}              
& \multicolumn{1}{|c|}{Disturbing Object}
& \textbf{num.=1}   
& \textbf{num.=2}   
& \textbf{num.=3}   \\ 
\midrule
\multirow{2}{*}{\begin{tabular}[c]{@{}c@{}} Cup \\ Insertion  \end{tabular}}      
                          & \multicolumn{1}{|c|}{\textbf{close-proximity}}          &    40  &  50    & 50 \\
                          & \multicolumn{1}{|c|}{\textbf{distant-proximity}}          &    70   &  70    & 60 \\
\bottomrule
\end{tabular}
\end{table}

\section{Conclusion} 
\label{sec:conclusion}

In this paper, we present Bimanual Keypose-conditioned Coordination-aware Consistency Policy (BiKC+), which is proposed for bimanual multi-stage manipulation tasks.
The focus of BiKC+ is to enhance the operational reliability and efficiency by effectively addressing the per-stage failures and per-step latencies accumulated during execution. 
BiKC+ is designed as a hierarchical imitation learning framework. It comprises a high-level keypose predictor that functions as a sub-goal planner and a low-level trajectory generator that serves as a behavioral policy.
The predicted keypose serves as the boundary between task sub-stages, offering guidance for generating actions to achieve sub-stage goals and acting as an indicator of sub-task completion.
The trajectory generator is formulated as a consistency model and is trained from scratch. It generates an action sequence conditioned on historical observations and target keypose through a single-step inference, thereby enhancing the inference speed. 
Furthermore, we propose an innovative approach to identify bimanual keyposes. This method leverages a vision-language model (VLM)-assisted contact-aware solution to enable automatic understanding of bimanual operation modes. Building on this foundation, bimanual keyposes are constructed via a coordination-driven merging strategy. These identified keyposes ensure that the BiKC+ policy achieves dual-arm synchronization in coordination modes while preserving the independence of individual arms during non-coordination phases, thereby enabling efficient and flexible bimanual robotic manipulation.
We evaluated BiKC+ in simulated and real-world tasks requiring bimanual coordination across multiple stages, covering both rigid and deformable objects.  Results demonstrated BiKC+'s superiority in success rates and operational efficiency, as well as multi-modality modeling capability.

\section{Discussion: Limitations and Future Works} 
\label{sec:discuss}

BiKC+ has yielded promising results in bimanual multi-stage manipulation tasks, while there remains potential for further exploration and improvement in the following perspectives.

\subsection{Bimanual Keyposes Identification} 
The first stage of keypose pipeline relies on heuristics for identifying unimanual keyposes. While these heuristics are capable of managing a wide range of tasks, there are certain situations that fall outside their applicability. 
A typical challenges occur when the boundaries of sub-tasks are not clearly delineated, as observed in periodic tasks such as turning a faucet and unscrewing a bottle cap. In these scenarios, it is difficult to accurately represent the outline of a task with a limited number of keyposes, as cyclic trajectories cannot be effectively approximated using a finite set of intermediate subgoals. Future research should focus on investigating these challenging scenarios in greater depth. 

Additionally, in the second stage of keypose pipeline, the identification of bimanual coordination mode depends on a VLM using RGB inputs. However, single-view RGB inputs inherently limit contact recognition accuracy, necessitating multi-camera integration to enhance VLM performance.
Specifically, the VLM determines contact as established if two objects show visual overlap from the camera’s perspective. This mechanism causes inaccuracies when the camera viewpoint is suboptimal. For example, objects may appear overlapping from a top-down view but are spatially separated from a front-facing angle, leading the VLM to false contact detection.
To address this, multi-camera setups are essential for the VLM-assisted contact-aware solution: they enable the VLM to cross-validate contacts across viewpoints, resolving viewpoint ambiguities and improving detection accuracy.
Currently, four static cameras are deployed (including one top camera, one front camera, and two wrist-mounted cameras) to maximize visual coverage. Nevertheless, their lack of active vision limits adaptability to dynamic scenes or complex scenarios. Future work can integrate an active vision control module \cite{chuang2025active} to optimize camera perspectives in real time, providing more accurate visual input to the VLM and further enhancing contact detection precision.

\subsection{Generative Model based Policy} 
In this paper, while the consistency model excels in approximating complex action sequence distributions and facilitating fast inference, certain limitations remain. For example, in the domain of image synthesis, CMs often struggle with rendering fine details such as fur and fingers. Similarly, in the field of robotic manipulation, they may fail to accurately reproduce the fine-grained trajectories with subtle variations in visual observations, which can lead to failures in tasks requiring ultra-high precision. 
Thus, it is essential to achieve a better balance between generation quality and speed. 
In the future, we intend to investigate advanced generative modeling techniques, such as diffusion models with sophisticated solvers~\cite{zheng2023dpm} and flow matching methods~\cite{chisari2024learning}.

Furthermore, according to the results of bimanual multi-stage experiments, BiKC+ and the baseline methods (i.e., ACT \cite{zhao2023learning} and DP \cite{chi2023diffusion}) encounter difficulties in specific stages, such as the ``insert" and ``press" phases in the Screwdriver Packing tasks. 
This is because reproducing subtle motions is challenging when only visual and positional sensors are available.
For example, the absence of force or tactile sensory input limits robot policies' ability to determine whether the screwdriver has been successfully inserted into the box, as seen in ACT and DP, or whether the grippers have adequately closed the lid, as observed in CP and BiKC+. 
To address this challenge, future research should integrate richer sensory information, such as data from force or tactile sensors(e.g., ~\cite{huang2024dvitac}), to enable accurate generation of subtle motions.

\appendix
\section{Appendix} 

\subsection{Recap of the Designs of Consistency Models} 
\label{sec:cm_details}

Several improved techniques are proposed in \cite{song2024improved} to enhance the performance of consistency models (CM) trained from scratch. We cover the new EMA decay rate, weighting function and loss function in Section~\ref{sec:preliminary} and we recap the remaining parts utilized in this paper here, including discretization, noise schedule, as well as the parameterization formulation.

\paragraph{Discretization curriculum} The continuous noise level $[\eps, \sigma_{\max}]$ is discretized into $N(k)$ points at the $k$-th training iteration, with $K$ iterations in total and $N(k)$ growing as $k$ increases. The new curriculum adopts a step-shaped curriculum where $N(k)$ doubles after a fixed number of iterations. Specifically,
\begin{equation}
\begin{aligned}
\label{eq:discretization_curriculum}
N(k)&=\min(s_0 2^{\lfloor \frac{k}{K'}\rfloor}, s_1)+1, \ 
K'&=\left\lfloor \frac{K}{\log_2\lfloor s_1/s_0 \rfloor + 1} \right\rfloor, \notag
\end{aligned}
\end{equation}
where $s_0=10,s_1=160$ are start and end numbers. Following~\cite{karras2022elucidating}, the discretized noises are given by $\sigma_i=(\eps^{1/\rho} + \frac{i-1}{N(k)-1}({\sigma_{\max}^{1/\rho}} - \eps^{1/\rho}))^\rho$, where $i\in\{1,2,\cdots,N(k)\},\rho=7,\eps=0.002,\sigma_{\max}=80$.

\paragraph{Noise schedule} Although \cite{song2024improved} proposes to sample $\sigma_i$ according to a lognormal distribution from the discretized sequence, we find that the original uniform distribution sampling performs better in our tasks. Therefore we stick to the old choice where $i\sim\gU\{1,2,\cdots,N(k)-1\}$.

\paragraph{Parameterization} In order to satisfy the self-consistency condition mentioned in Section~\ref{sec:preliminary}, authors of CM~\cite{song2023consistency} leverage a special parameterization to make $\vf_{\bm{\theta}}(\rvx_\eps,\eps)=\rvx_\eps$ hold by design:
\begin{equation}
\vf_{\bm{\theta}}(\rvx,\sigma) \coloneqq c_{\text{skip}}(\sigma)\rvx + c_{\text{out}} \mF_{\bm{\theta}}(\rvx,\sigma), \notag
\end{equation}
where $c_{\text{skip}}(\sigma) = {\sigma_\text{data}^2}/((\sigma-\eps)^2+\sigma_\text{data}^2)$ and $c_{\text{out}}(\sigma) = \sigma_\text{data}(\sigma-\eps)/\sqrt{\sigma_\text{data}^2+\sigma^2}$ satisfies $c_{\text{skip}}(\eps)=1$ and $c_{\text{out}}(\eps)=0$ and $\mF_{\bm{\theta}}$ is the parameterized model (e.g., a neural network).

\subsection{Example of Contact-Relationship Identification with VLM} 

When implementing a VLM-assisted contact-aware solution, we extract the image frames at 10Hz from the demonstration video and take the sequence of image frames as input to VLM model. Here, we present the PROMPT to the VLM to detect changes in object relationships. \\

\promtblock{\underline{\textbf{Prompt:}}\\
\footnotesize
\newline
Task: $\{\text{\$TASK\_NAME} \}$ \\
Task-related objects: $\{\text{\$OBJECT\_LIST} \}$ \\
Supporting surfaces:  $\{\text{\$SURFACE\_LIST} \}$ \\
Video duration: $\{\text{\$TIME\_LENGTH} \}$ seconds \\
\newline
You will analyze a sequence of video frames to detect contact relationship changes, given the information below: \\
\newline
\textit{FRAME INDEXING GUIDE:} \\
The files are named as \{ frame\_00000.jpg, frame\_00001.jpg, $\cdots$ \}. The frame index \texttt{<frame\_idx>} corresponds to the value in the filename. \\
\newline
\textit{CRITICAL CONTACT DETECTION RULES:}
\begin{enumerate}
    \item \textbf{Grasp (robot-object):} Only when robot gripper/fingers are PHYSICALLY CLOSED around the object AND the object moves with the robot
    \item \textbf{Approcah vs Grasp:} Robot moving toward object $\neq$ grasping. Only report grasp when contact is ESTABLISHED and MAINTAINED
    \item \textbf{Visual Confirmation:} Look for:
    \begin{itemize}
        \item Both of gripper fingers touching/enclosing the object
        \item Object displacement/movement caused by robot action
        \item Clear physical contact, not just proximity
    \end{itemize}
\end{enumerate} 
$\quad$ \\
\textit{Answer the following questions step by step based on the given frame sequence:}
\begin{enumerate}
    \item Using objects, robots, and surfaces as nodes, and the contact relationship among them as edges, construct a graph to describe the initial contact relationship among them. The \texttt{<initial\_graph>} is made up of a list of edges and nodes of the whole scene. Initially, all objects should connect to the surface.
    \item Starting from \texttt{<initial\_graph>}, detect contact mode changes by analyzing each video frame in sequence.
    Only report contact changes when you can CLEARLY observe:
    \begin{itemize}
        \item Physical contact establishment or breaking
        \item Object state changes (lifted, moved, released)
    \end{itemize}
    Establishing contact between entities corresponds to an \textbf{Add} operation, while breaking contact corresponds to a \textbf{Remove} operation.
\end{enumerate}
$\quad$ \\
\textit{Provide the final output in the following format, without additional explanation:}\\
\newline
$\quad$ 
\textit{Initial Graph}: \texttt{<initial\_graph>} \\
$\quad$ 
\textit{ModeChangeDetection}: [\{\texttt{<frame\_idx>}, \texttt{<edge>}, \\
\texttt{<Add> or <Remove>}) ]  \\
\newline
\llmoutput{\underline{\textbf{VLM output:}} (\textit{Transfer Cube} as Example) \\
\footnotesize
\newline
\textit{\textbf{Initial Graph}}: [(`cube', table'),(`robot left', `table'),(`robot right', `table' )] \\
\textit{\textbf{ModeChangeDetection}}: [(15, (`robot right', 'cube'), `Add'), (40, (`table', `cube'), `Remove'), (60, (`robot left' ,`cube') ,`Add'), (70, (`robot right' ,`cube') ,`Remove'), (80, (`table', `cube'), `Add'))] \\
}}

\subsection{Detailed Hyperparameters} 
\label{sec:hyper_params}

We generally set hyperparameters to the values suggested in the original articles for all baselines and include them in Tables~\ref{tab:hyper_ACT} and~\ref{tab:hyper_diffusion}.

\begin{table}[!h]
\centering
\caption{Hyperparameters of ACT in real-world tasks.}
\label{tab:hyper_ACT}
\begin{tabular}{llll}
\toprule
\multicolumn{1}{c}{Parameter} & \multicolumn{1}{c}{Value} & \multicolumn{1}{c}{Parameter} & \multicolumn{1}{c}{Value} \\ \midrule
Learning rate            & 1e-5                      & Num. of encode layers         & 4                         \\
Batch size               & 8                         & Num. of decoder layers        & 7                         \\
Chunk size               & 100                       & Num. of heads                 & 8                         \\
KL weights               & 10                        & Feedforward dim         & 3200                      \\
Num. of epochs           & 10000                     & Hidden dim              & 512                       \\ \bottomrule
\end{tabular}
\vspace{-4mm}
\end{table}

\begin{table}[htb]
\centering
\caption{Hyperparameters of Diffusion-based Algorithms on simulated and real-world tasks, where some values are in the form of [simulation, real-world].}
\label{tab:hyper_diffusion}
\begin{tabular}{@{}ll@{}}
\toprule
    Parameter & Value \\
\midrule
    \emph{Shared} & \\
    \quad Learning rate           & 1e-4 $\rightarrow$ 0 \\
    \quad Weight decay            & 1e-6 \\
    \quad Batch size              & 64\\
    \quad Num. of epochs               & 500, 1000\\
    \quad ImgRes                  & 480$\times$640, 4$\times$480$\times$640\\
    \quad DownRes                 & 240$\times$320, 4$\times$120$\times$160\\
    \quad CropRes                 & 220$\times$300, 4$\times$110$\times$150\\
    \quad Proprio dim             & 14 \\
    \quad Sub-trajectory horizon  & 16\\
    \quad Obs horizon             & 2\\
    \quad Action horizon          & 8\\
    \quad Model arch              & 1D Unet\\
    \quad Diffusion step emb dim  & 128\\
    \quad Keypose emb dim         & 128\\
    \quad Unet conv channel dim   & 256-512-1024\\
\midrule
    \emph{DP} &\\
    \quad Num. of train diffusion steps & 100\\
    \quad Num. of eval diffusion steps  & 10,16\\
    \quad EMA decay rate           &  0.75\\
\midrule
    \emph{CM} &\\
    \quad $\sigma_{\text{data}}$   & 0.5\\
    \quad EMA decay rate           & 0\\
\bottomrule
\end{tabular}
\vspace{-4mm}
\end{table}

\bibliographystyle{IEEEtran}
\bibliography{IEEEabrv, reference}


\begin{IEEEbiography}
[{\includegraphics[width=1in,height=1.25in,clip,keepaspectratio]{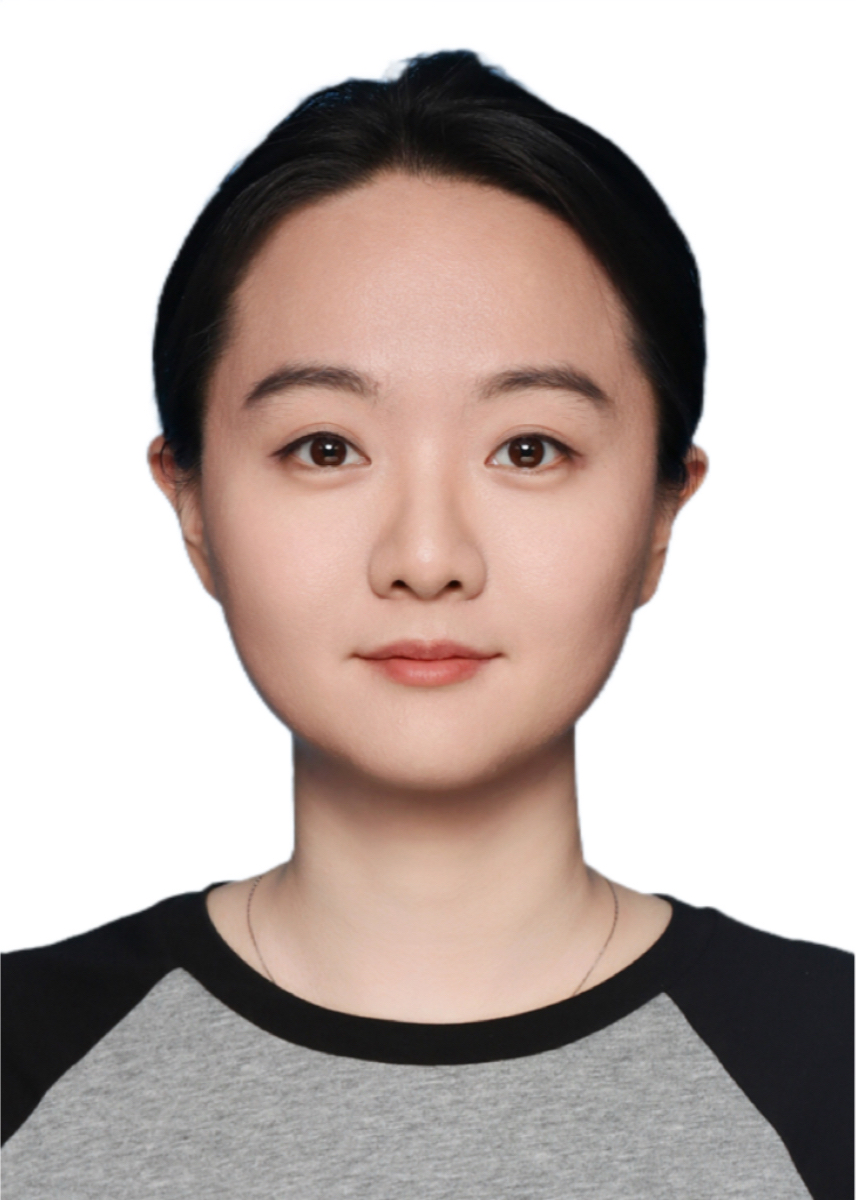} \hspace{-2pt}}] {Hang Xu} received her B.S. degree in communication engineering from Wuhan University, China, in 2015; her M.S. degree from the School of Electrical and Electronics Engineering, Nanyang Technological University, Singapore, in 2017; and her Ph.D. degree from the School of Computer Science and Engineering, Nanyang Technological University, Singapore, in 2023.

She is currently a Research Officer in the Centre for Transformative Garment Production, The University of Hong Kong, Hong Kong. 
Her recent research interests focus on the intersection between dexterous robotic manipulation and machine learning algorithms, including imitation learning and reinforcement learning.
\end{IEEEbiography}
%
\begin{IEEEbiography}
[{\includegraphics[width=1in,height=1.25in,clip,keepaspectratio]{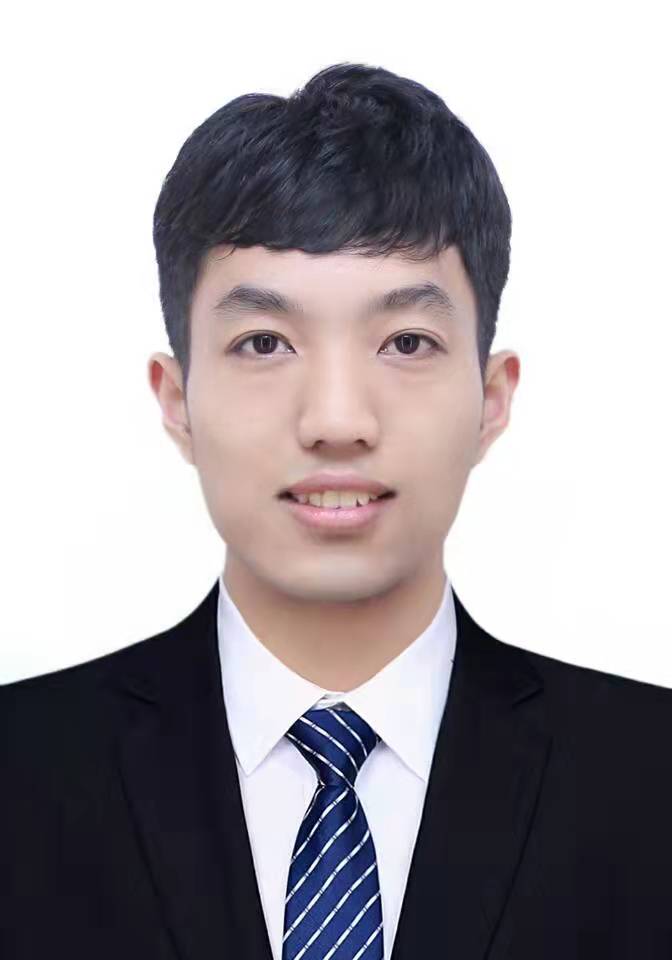}}]{Yizhou Chen} received his B.S. degree in automation from Nanjing University of Aeronautics and Astronautics, Nanjing, China, in 2019; and his Ph.D. degree in mechanical and automation engineering from the Chinese University of Hong Kong, Hong Kong, in 2023. From 2023 to 2025, he was a Postdoctoral Fellow with Center for Transformative Garment Production, The University of Hong Kong, Hong Kong, China.

He is currently a Postdoctoral Fellow in the Department of Computer Science, The University of Hong Kong, Hong Kong.
His current research interests include parallel computing, formal methods, and task and motion planning.
\end{IEEEbiography}
%
\begin{IEEEbiography}
[{\includegraphics[width=1in,height=1.25in,clip,keepaspectratio]{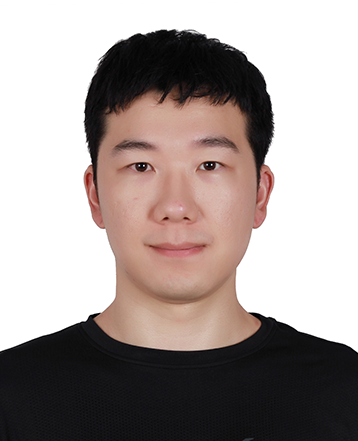}}]{Dongjie Yu} received his B.S. and M.S. degrees in Mechanical Engineering at the School of Vehicle and Mobility from Tsinghua University, Beijing, China, in 2020 and 2023, respectively.

He is currently pursuing his Ph.D. degree in the Department of Computer Science, The University of Hong Kong, Hong Kong. 
His current research interests include robot learning approaches and their application in robotic manipulation.
\end{IEEEbiography}

\begin{IEEEbiography}
[{\includegraphics[width=1in,height=1.25in,clip,keepaspectratio]{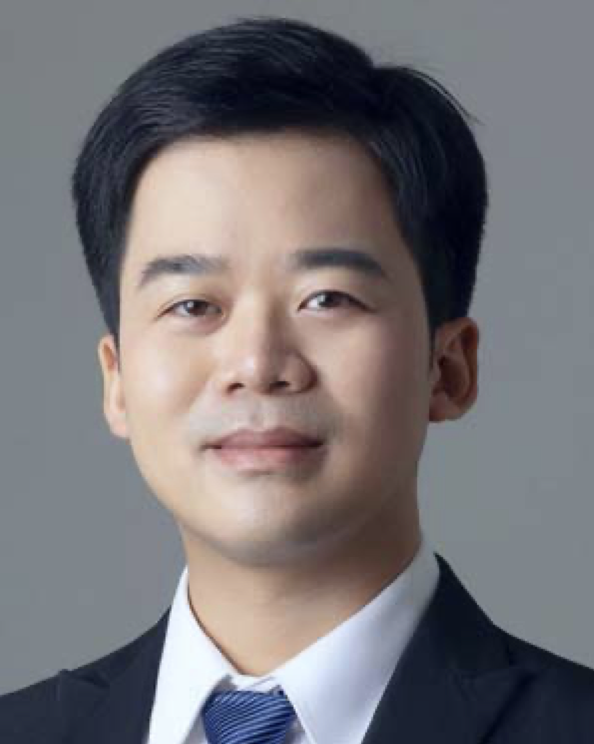}}]{Yi Ren} (Member, IEEE) received the B.S. degree in thermal energy and power engineering from Harbin Institute of Technology, Harbin, China, in 2011, and the M.Sc. and Ph.D. degrees in mechatronics engineering from the State Key Laboratory of Robotics and System, Harbin Institute of Technology, in 2013 and 2017, respectively.
From 2018 to 2020, he was a Postdoctoral Fellow and Senior Scientist in electrical engineering with the Chair of Information-oriented Control, Department of Electrical and Computer Engineering, Technical University of Munich, Munich, Germany. From 2020 to 2022, he was a Research Scientist with Tencent Robotics X Lab. 

He joined the Advanced Manufacturing Lab, Huawei Technologies, Shenzhen, China, in 2023, and is currently a technical expert in Robotics. His research interests include robotics, nonlinear control, distributed control, perception, and natural language processing with application to multirobot cooperation and manipulation.
\end{IEEEbiography}

\begin{IEEEbiography}
[{\includegraphics[width=1in,height=1.25in,clip,keepaspectratio]{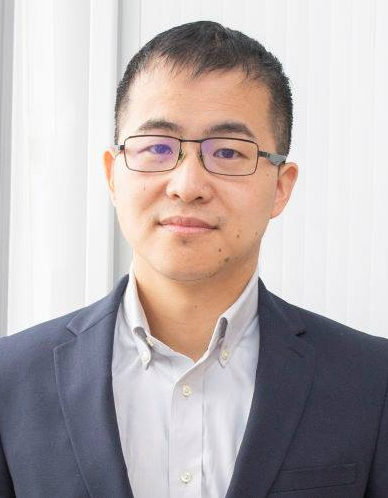}}]{Jia Pan} (Member, IEEE) received the B.S. in Control Theory and Engineering from Tsinghua University, China, in 2005; his M.S. at the National Laboratory of Pattern Recoginition, Institute of Automation, Chinese Academy of Science in 2008; and his Ph.D. in the Department of Computer Sciencee at the University of North Carolina at Chapel Hill, USA, in 2013.

He is currently an Associate Professor in the Department of Computer Science at The University of Hong Kong. His research focuses on robotics and AI for autonomous systems, including crowd navigation, deformable object manipulation, and construction robotics. 
\end{IEEEbiography}

\end{document}